\documentclass[acmsmall]{acmart}
\AtBeginDocument{%
  \providecommand\BibTeX{{%
    \normalfont B\kern-0.5em{\scshape i\kern-0.25em b}\kern-0.8em\TeX}}}

\setcopyright{acmcopyright}
\copyrightyear{2018}
\acmYear{2018}
\acmDOI{10.1145/1122445.1122456}

\acmJournal{JACM}
\acmVolume{37}
\acmNumber{4}
\acmArticle{111}
\acmMonth{8}

\usepackage{amsmath}
\usepackage{multirow}
\usepackage{array,bm}

\begin{document}

\title{Identifying Illicit Drug Dealers on Instagram with Large-scale  Multimodal Data Fusion}

\author{Chuanbo Hu}
%\authornote{This work is partially supported by the NSF under grants IIS-2107172, IIS-2027127, IIS-2040144, CNS-2034470, IIS-1951504, CNS-1940859, CNS-1814825 and OAC-1940855, and the DoJ/NIJ under grant NIJ 2018-75-CX-0032.   }
\email{cbhu@whu.edu.cn}
\orcid{1234-5678-9012}
\author{Minglei Yin}
\authornotemark[1]
\email{my0033@mix.wvu.edu}
\author{Bin Liu}
\email{bin.liu1@mail.wvu.edu}
\author{Xin Li}
\authornotemark[1]
\email{xin.li@mail.wvu.edu}
\affiliation{%
  \institution{West Virginia Univesity}
  \streetaddress{P.O. Box 6109}
  \city{Morgantown}
  \state{WV}
  \postcode{26506-6109}
}

\author{Yanfang Ye}
\affiliation{%
  \institution{Case Western Reserve University}
  \streetaddress{Dept. of CDS}
  \city{Cleveland}
  \state{Ohio}
  \postcode{78229}}
\email{yanfang.ye@case.edu}

\renewcommand{\shortauthors}{Hu and Yin, et al.}

\begin{abstract}
Illicit drug trafficking via social media sites such as Instagram has become a severe problem, thus drawing a great deal of attention from law enforcement and public health agencies. How to identify illicit drug dealers from social media data has remained a technical challenge due to the following reasons. On the one hand, the available data are limited because of privacy concerns with crawling social media sites; on the other hand, the diversity of drug dealing patterns makes it difficult to reliably distinguish drug dealers from common drug users. Unlike existing methods that focus on posting-based detection, we propose to tackle the problem of \emph{illicit drug dealer identification} by constructing a large-scale multimodal dataset named Identifying Drug Dealers on Instagram (IDDIG). Totally nearly 4,000 user accounts, of which over 1,400 are drug dealers, have been collected from Instagram with multiple data sources including post comments, post images, homepage bio, and homepage images. We then design a quadruple-based multimodal fusion method to combine the multiple data sources associated with each user account for drug dealer identification. Experimental results on the constructed IDDIG dataset demonstrate the effectiveness of the proposed method in identifying drug dealers (almost 95\% accuracy). Moreover, we have developed a hashtag-based community detection technique for discovering evolving patterns, especially those related to geography and drug types.
\end{abstract}

\begin{CCSXML}
<ccs2012>
   <concept>
       <concept_id>10002951.10003227.10003351</concept_id>
       <concept_desc>Information systems~Data mining</concept_desc>
       <concept_significance>500</concept_significance>
       </concept>
   <concept>
       <concept_id>10010405.10010455</concept_id>
       <concept_desc>Applied computing~Law, social and behavioral sciences</concept_desc>
       <concept_significance>500</concept_significance>
       </concept>
 </ccs2012>
\end{CCSXML}

\ccsdesc[500]{Information systems~Data mining}
\ccsdesc[500]{Applied computing~Law, social and behavioral sciences}

\keywords{drug trafficking; drug dealer; Instagram; multimodel data fusion}

\maketitle

\section{Introduction}

% illicit drug trade is a serious problem
Illegal drug trade (a.k.a. drug trafficking) is a global black market dedicated to the distribution and trade of drugs prohibited by law. Due to the co-evolution of cyberspace and human society, online illicit drug trade has become a major problem, attracting increasingly more attention from both law enforcement and public health agencies. According to a RAND report in 2010 \cite{kilmer2014big}, the total size of the illicit drug market in the US was estimated to be in the order of 100 billion dollars per year. Recent studies \cite{yang2017tracking,li2019machine} have shown that popular social media platforms such as Instagram, Twitter, and Facebook have become a convenient direct-to-consumer marking tool for illegal dealers. Among popular social media platforms, Instagram is a particularly effective tool for advertising illicit drugs due to their photo-sharing features. Since the majority of Instagram users are young people, including teenagers, it is important to tackle the problem of illicit drug trade on Instagram.

\begin{figure*}
\begin{center}
   \includegraphics[width=0.95\textwidth]{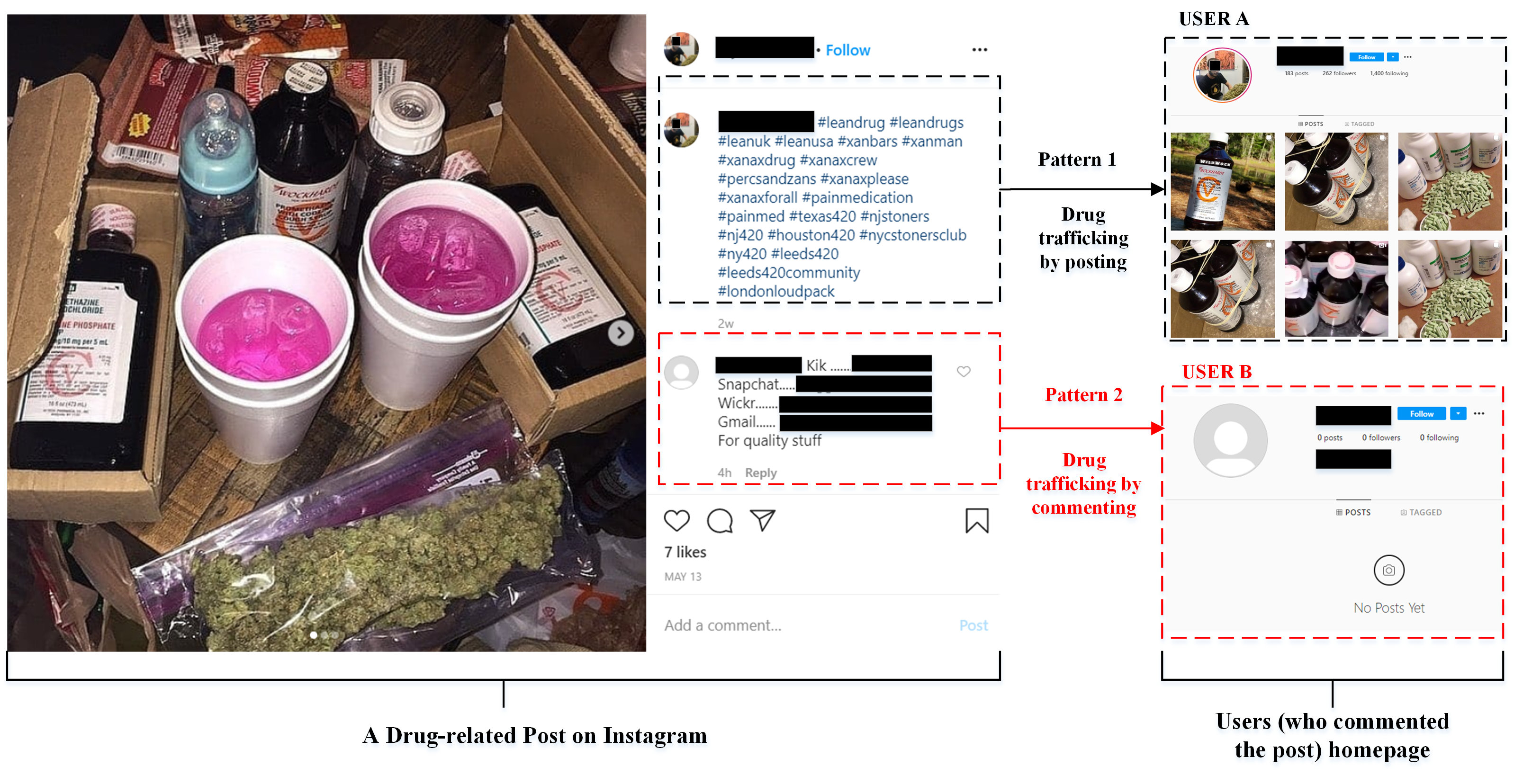} 
\end{center}
  \caption{A Drug-related Post with Multiple Patterns of Illicit Drug Dealing. Drug dealers can engage in illicit drug trade by direct posting (pattern 1) or commenting (pattern 2).}
  \Description{An example of illicit drug dealing.}
  \label{fig:1}
\end{figure*}

% Technical challenges facing illicit drug dealer detection
In this paper, focusing on Instagram, we study the problem of  \emph{illicit drug dealer identification}, which is a critical step to combat  illicit drug trafficking on social media platforms. 
There are several technical challenges facing the detection of illicit drug dealers from Instagram data: (1) Ambiguity. It is often difficult to distinguish an illicit drug dealer from a drug abuser account. Even for domain experts, different people could reach different conclusions, which make the results inconsistent \cite{yang2017tracking}.  
Illicit drug trafficking methods have become sophisticated in social media platforms such as Instagram. 
As shown in Fig. \ref{fig:1}, drug dealers can engage in illicit drug trade by either posting or commenting. Furthermore, how to distinguish illicit drug dealers from legal regular consumers or drug abuser accounts has remained one of the open problems \cite{yang2017tracking}. (2) Heterogeneity. The data sources related to illicit drug trade involve both images and text; moreover, the modes of communication adopted by illicit drug dealers are diverse - posting vs. commenting, hashtag vs. homepage, and diversity of drug-related images. How to systematically pull together these heterogeneous information has remained a long-standing open problem \cite{sun2012mining}. (3) Scalability. Even though the population of drug dealer accounts is relatively small, the total size of social media data is enormous. How to efficiently mine large-scale and temporally varying social media data calls for innovative technical solutions at the system level \cite{zafarani2014social}.

% New insight provided by this work
The motivation behind this work is largely two-fold. First,   
recognizing the importance of data and the rapid growth of Instagram community \cite{greenwood2016social}, we argue that it is necessary to construct an update-to-date and large-scale database to support the research related to illicit drug trade. When compared against the pioneering work in this field \cite{yang2017tracking}, we believe Instagram-based detection of illicit drug dealers can benefit from a more ambitious effort of data collection. Moreover, we advocate the public release of such a database to serve as a benchmark for supporting the emerging field of research on illicit supply networks \cite{hughes2017social}. Second, rapid advances in machine learning, especially deep learning \cite{goodfellow2016deep} and multimodal data fusion \cite{lahat2015multimodal} provide a rich weaponry of advanced computational tools for automatically extracting knowledge from training data without feature engineering. How to leverage these latest advances in drug dealer identification calls for an integrated approach. Meanwhile, the issue of dealing with missing information in multimodal data (e.g., drug dealers often intentionally hide some information for evading law enforcement) has remained an open problem \cite{lahat2015multimodal}.

% What is this work about? 
In this work, we propose to tackle the problem of drug dealer identification by constructing a large-scale dataset called Identifying Drug Dealers on Instagram (IDDIG), including over 2,000 posts, nearly 4,000 user homepages as well as multimodal data sources (text and images, posts and homepage). In particular, the importance of biography information at the homepage on social media data mining has remained an underexplored topic \cite{hu2014we,jang2015generation} and not been studied in previous works on drug-dealer detection \cite{yang2017tracking,li2019machine}. To construct such a large-scale dataset, we have designed an automatic data crawling system for Instagram that jointly uses hashtag and image information to guide the data collection. A user-friendly data annotation platform is also developed to support manual labeling of multimodal data, which is inevitably a tedious process.  With labeled data (ground-truth for machine learning), we have built a brand new drug-dealer identification system by leveraging the latest advances in deep learning including Bidirectional Encoder Representations from Transformers (BERT) \cite{devlin2018bert} based text classification,  ResNet-based \cite{he2016deep} image classification, as well as feature-level multimodal data fusion \cite{gao2020survey}. %Fig. \ref{fig:4} shows the overview of the developed fully automatic $iDetector$ system for illicit drug dealer identification from Instagram data.
% Summary of our contributions
The key contributions of this paper are summarized as follows. 
\begin{itemize}
    \item The construction of a large-scale dataset called IDDIG for drug-dealer identification. We have designed an automatic hashtag-based data crawling system and a user-friendly data annotation system to support large-scale data collection. For the first time, we have collected over 1,400 drug dealers' accounts (positive examples) and over 2,000 non-drug-dealer accounts (negative examples). The newly constructed IDDIG dataset will be made publicly available to support the research related to illicit drug trade.
    
    \item The development of a fully automatic and highly accurate system for drug dealer identification. Our new system is built upon a novel quadruple-based data representation (image vs. text, post vs. homepage) and several latest advances in deep learning including natural language processing (e.g., BERT model \cite{devlin2018bert}), deep image classification (e.g., ResNet model \cite{he2016deep}) and feature-level multimodal data fusion (e.g., \cite{haghighat2016discriminant}). For the first time, we have addressed the issue of missing modalities in multimodal data fusion. %The identification system built upon quadruple-based fusion has achieved an overall accuracy of nearly 95\%. 
    
    \item The application of joint post and homepage identification to real-world Instagram data for identifying illicit drug dealers and drug rings (i.e., community detection). We have achieved an overall accuracy of almost $95\%$ on our test dataset with a balanced training dataset, and demonstrated a graceful performance degradation as the training dataset gradually becomes unbalanced. The developed system has also demonstrated its potential for detecting drug-trafficking communities online, which could facilitate the disruption of illicit drug trade by law enforcement.
\end{itemize}

The rest of this paper is organized as follows. In Section \ref{sec:2}, we briefly review related works on social media data mining and user profiling to contextualize the proposed approach. In Section \ref{sec:3}, we describe the construction of our new large-scale dataset in detail, including data crawling and data annotation subsystems. In Section \ref{sec:4}, we present a new drug-dealer identification system by jointly exploiting the post and homepage information. In Section \ref{sec:4a}, we address the issue of community detection based on hashtags and graph clustering. We report our experimental results in Sec. \ref{sec:5} and make several concluding remarks in Sec. \ref{sec:6}.

\section{Related Work}
\label{sec:2}

\subsection{Social Media Data Mining}

In the past decades, social media has greatly facilitated the generation and sharing of information via virtual communities and networks. Data associated with popular social media platforms such as Facebook and Twitter have grown at exponential rates. Accordingly, social media data mining \cite{barbier2011data, zafarani2014social,song2015multiple,song2015interest,song2016volunteerism} has rapidly evolved into an emerging field with a wide range of applications from online surveys \cite{oConnor2010tweets} and public health monitoring \cite{thackeray2012adoption,luxton2012social} to online marketing \cite{scott2015new, nie2016learning} and recommendation systems \cite{guy2010social,ling2014ratings}. Various new mining techniques such as user profiling and community detection have been developed for social media data specifically.

In particular, our work is related to  user profiling in social media that classifies users into different labels  from their social media data such as posted texts and images \cite{li2012towards,ikeda2013twitter}. Such a collection of online identities has found successful applications in personalized recommendation \cite{hung2008tag} and marketing analysis \cite{ikeda2013twitter}. Most recently, deep learning-based multimodal data fusion has been studied for user profiling in \cite{farnadi2018user}. Conceptually analogous to the grouping of people in the physical world, online identities tend to form various virtual communities. Community detection and analysis \cite{barbier2011data, zafarani2014social} have been widely studied in the literature of social media data mining. Both member-based and group-based community detection algorithms have been developed; both the evolution and evaluation of virtual communities have been investigated. 

\subsection{Drug Abuse and Dealing Analysis}
%\subsection{Existing Algorithms}
As far as we know, there has been limited work on tracking drug abuse and illicit drug trade from online data. Among these existing works, \cite{buntain2015your} analyzed the time and
location patterns of drug use by mining Twitter data; network information of Instagram user timelines was used in \cite{correia2016monitoring} to monitor suspicious drug interaction activities; \cite{zhou2016understanding} and \cite{yang2017tracking} analyzed Instagram data for tracking and identifying drug dealer accounts. More recently, machine learning and natural language processing techniques have been applied to combat prescription drug abuse \cite{kalyanam2017review,sarker2020mining,sarker2019machine} and detect illicit drug dealers \cite{li2019machine,zhao2020computational}. Different from previous work that only used one modality data (i.e., text data), our work is built upon multimodal data fusion for illicit drug dealer identification. There are some works that combines both image and text data for identifying substance use risk \cite{hassanpour2019identifying}. Our work is different in twofold: first, our research goal is illicit drug dealer identification, which is more challenging due to the factor that illicit drug trafficking methods have become sophisticated in social media platforms. Second, our  multimodal data fusion is built upon a novel quadruple-based data representation (image vs. text, post vs. homepage). 

\subsection{Illicit Drug Use Detection Systems}
In our own previous work, we have developed an automatic opioid user detection system called $AutoDOA$ using Twitter data in \cite{fan2017social}. In $AutoDOA$, to model the users and posted tweets as well as their rich relationships, we first construct a structured heterogeneous information network (HIN) \cite{shi2016survey}. Then we have taken a meta-path based approach to formulate similarity measures over users and aggregate different similarities using Laplacian scores. This work was further extended in \cite{fan2018automatic} into $HinOPU$ by constructing an ensemble of classifiers to combine different predictions, which improves the accuracy for opioid user detection. Most recently, we have developed a novel and intelligent system named $uStyle$-$uID$ leveraging both writing and photography styles for drug trafficker identification on darkweb \cite{zhang2019your}. At the core of this new system is attributed heterogeneous information network (AHIN) \cite{li2017semi} which elegantly integrates both writing and photography styles along with the text and photo contents, as well as other supporting attributes (i.e., drug dealer and drug product information) and various kinds of relations. However, it is usually hard to detect illicit drug trafficking activities from Instagram data based on these existing methods due to the following reasons. First, Instagram is an image-oriented social media platform, which is different from text-oriented ones such as Twitter. Therefore, it is difficult to directly transfer $AutoDOA$ \cite{fan2017social} from Twitter to Instagram due to the variation of modality. Second, $uStyle$-$uID$ \cite{zhang2019your} is developed for darkweb, whose characteristics differ dramatically from those of clearweb - e.g., it is generally difficult for the public to access darkweb; but once it is accessed, it is relatively easy to detect illicit drug trafficking. By contrast, detecting illicit drug trafficking on clearweb is like finding a needle in a haystack. Most existing Instagram-based detection methods can not comprehensively detect constantly evolving drug trafficking patterns or activities. For example, \cite{yang2017tracking, li2019machine} can detect illegal drug trafficking such as Pattern 1 in Fig. \ref{fig:1} (direct posting) with good performance but fail to detect Pattern 2 (indirect commenting). %More kinds of illegal drug trafficking activities can be detected by the proposed method. To solve this problem, we collect the IDDIG dataset and proposed a quadruple-based fusion method.}

\begin{table*}[]
\caption{Existing Drug-related Datasets from Social Media.}
\label{tab:my-table}
\newcommand{\tabincell}[2]{\begin{tabular}{@{}#1@{}}#2\end{tabular}}
\centering
\small
\begin{tabular}{ccccccc}
\hline 
Study & Year & Data source & Drug user/Drug dealer & \# of positive posts & \tabincell{l}{\# of positive \\ user accounts} & \tabincell{l}{Public \\ available} \\
\hline 
\cite{zhou2016fine} & 2016 & Instagram & Drug user & 2,362 & 406 & No \\
\cite{yang2017tracking} & 2017 & Instagram & Drug dealer & 1,260 & 27 & No \\
\cite{mackey2018solution} & 2018 & Twitter & Drug dealer & 692 & / & No \\
\cite{hu2019insight} & 2019 & Twitter & Drug user & 280 & / & No \\
\cite{li2019machine} & 2019 & Instagram & Drug user and dealer & 1,228 & 237 & No \\
IDDIG (Ours) & 2020 & Instagram & Drug dealer & 1,022 & 1,406 & Yes  \\
\hline 
\end{tabular}
\end{table*}

\section{IDDIG Dataset Construction}
\label{sec:3}

Data collection has played an increasingly important role in machine learning and data mining research \cite{stieglitz2018social}. In this section, we first review several existing datasets related to drug trafficking research. Then we present a new large-scale dataset constructed from Instagram; toward this objective, we will discuss our efforts on data crawling and data annotation, respectively.

\subsection{Current State-of-the-Art}

%{\color{red} Please provide a one-paragraph description of each dataset in this subsection and add the reference to Table 1.}

Existing detection methods have collected several databases for drug-dealer and drug-user detection. Tab. \ref{tab:my-table} summarizes the statistics of these datasets for illicit drug dealing tracking. In \cite{zhou2016fine}, a dataset containing 2,362 posts and 406 drug users (including drug dealers) was introduced. It also contained drug users’ followers and the accounts they follow. 176 users contain geolocation tags in their posts. Based on the post and homepage information on Instagram, a total of 1,260 drug-related posts and 27 drug dealer accounts were collected in \cite{yang2017tracking} for drug dealer detection. Each account in the dataset contains detailed information, including post content, temporal information of each post, and the number of followers/followees. In \cite {mackey2018solution}, 692 drug-related tweets were identified as being associated with illegal online marketing and sale of prescription opioids to detect and report illicit online marketing and sales of controlled substances via Twitter. A recent study \cite{hu2019insight} collected a dataset (with 280 drug-related) for analysis of drug‑abuse risk behavior on Twitter. Recently, 1,228 drug dealer posts and 267 drug dealer accounts were collected by \cite{li2019machine} and images of different types of drugs were fine-grained and categorized into five different categories.  

Although great effort has been devoted to drug-related data collection, there exist the following limitations of existing datasets: (1) The size of these datasets are not large enough to support deep learning-based approaches. For example, the number of drug dealer or user account instances is all less than 500. This is limited by that drug dealers or users make up a tiny fraction of Instagram's user accounts \cite{yang2017tracking}, and many heterogeneous data (e.g., text and images) need to be collected, increasing the cost of storage and management \cite{hassanpour2019identifying}. (2) Due to the privacy protection, the user account number, face related images and contact information (e.g., other chat app user ID, see the black blocks in Fig. \ref{fig:1}) are not allowed to be made public. Consequently, no dataset is publicly available to support the research related to drug dealer detection as of today.

\subsection{Multi-model Data Collection Scheme on Instagram}

To fill in this gap, we have made a great effort on data collection and construction of a large-scale dataset in this project. Existing web crawling techniques (e.g., \cite{olston2010web}) do not meet the requirements of this project due to the following technical challenges. First, it is desirable to develop an automatic system capable of identifying drug-related content without human intervention. This is in sharp contrast to previous works (e.g., \cite{yang2017tracking}) that count on humans for data collection, which are difficult to scale up. Second, the designed data crawling system needs to be versatile in that it supports multimodal (including both text and images) and multisource (e.g., posts and comments) collection. 

\begin{figure*}[t]
  \centering
  \includegraphics[width=0.95\linewidth]{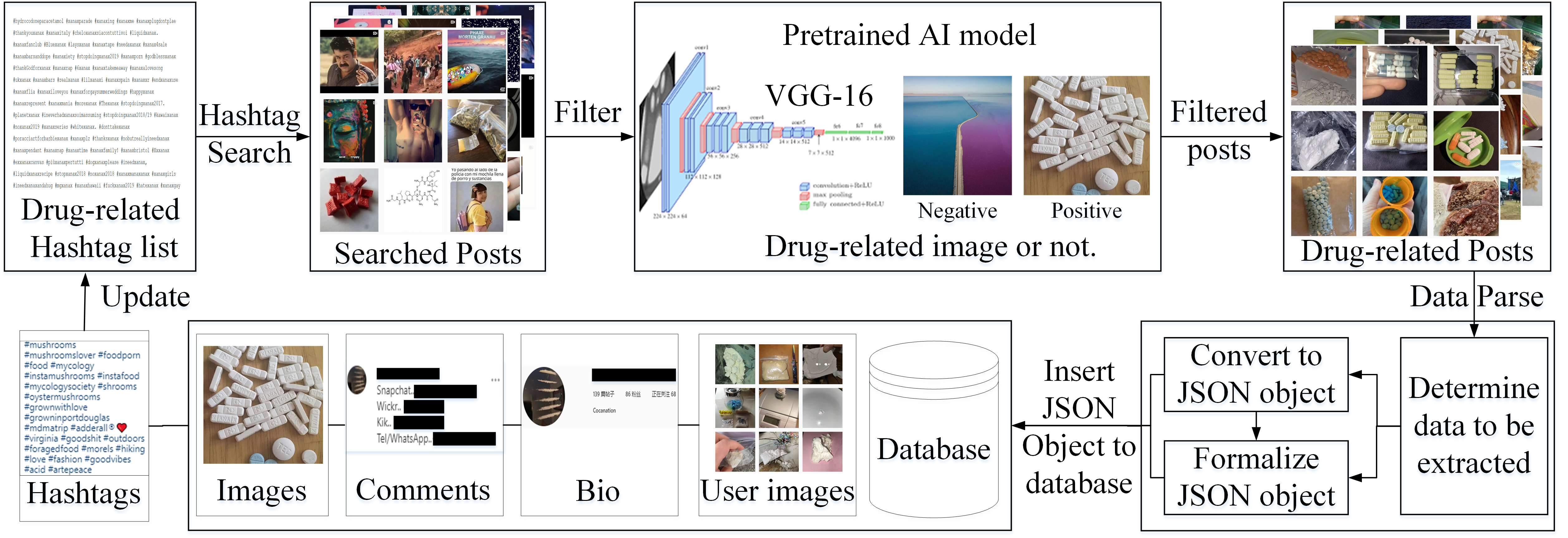}
  \caption{Automatic and Iterative Hashtag-based Multimodal Data Collection from Instagram.}
  \Description{}
  \label{fig:2}
\end{figure*}

To achieve this objective, we have developed an automatic system capable of iteratively collecting multimodal data, as shown in Fig. \ref{fig:2}. The rationale underlying our data crawling system is still based on {\it hashtag}-based search \cite{godin2013using}. Hashtags on Instagram can help users extend their reach, engage their audience, which can be attached to posts and become clickable phrases and topics with the $\#$ placed in front of them. However, unlike \cite{yang2017tracking} working with a fixed collection of hashtags, we propose a data crawling algorithm that iteratively expands the pool of hashtags for scaling up our search. Such expansion of hashtags is guided by an intelligent pretrained AI model (VGG-16 \cite{simonyan2014very}) designed for drug image classification. By treating drug-related hashtags and images as a pair of peer hidden variables, our iterative crawling system aims at refining and updating the collected multimodal data in an Expectation-Maximization (EM)-like manner. The detailed description of our data collection system consists of the following four components.

(1)	Drug-related hashtags collection. A total of 200 drug-related hashtags have been manually collected by domain experts using the hashtag search API \cite{gao2017hashtag}. These hashtags contain various types of drugs, such as 3,4-methylenedioxy-methamphetamine (MDMA), Lysergic acid diethylamide (LSD), codeine, painkiller and xanax etc., which are widely trafficked on Instagram. We have used this set of hashtags as the initial starting point of our data collection.

(2)	Drug-related post detection. We search each post (which includes an image and comments) with each drug-related hashtag as input. A VGG-16 based binary classification model \cite{simonyan2014very} is pre-trained to detect drug-related posts from the accompanying image information. The image-based dataset for model pretraining contains various types of drug-related images, which are sources of Bing image search API (similar to Google image search API adopted in \cite{yang2017tracking}). If an image of a post is detected by the model as being drug-related (positive), we save its link for further processing.

(3)	Drug-related data collection. The detected posts were converted and formalized into a universal Json object \cite{maeda2012performance} to facilitate storage and retrieval. As post comments are sources from several user accounts, we saved each post-related information (including posted images and comments) and homepage information about the user who commented on the post. The user’s homepage information includes both the bio and images from the user’s latest 10 posts (we refer to it as Homepage image in the following). Totally, 10,000 potential posts and 23,034 user homepage information were collected as the initial dataset. 

(4)	Drug-related hashtag update. New hashtags from each detected post can be added into the list of drug-related hashtags. We have also recorded the frequency of each hashtag to track the most frequent ones. The system uses the new hashtag (which has the highest frequent counts) in the next iteration until the amount of collected data reaches a prespecified threshold (in this study, we have set the threshold to be 1000 drug-dealer accounts). 

When compared with previous work \cite{yang2017tracking}, we note that the key difference lies in our emphasis on posted comments and biography data (Instagram bio) instead of hashtags and captions as salient features for drug-dealer identification.

\subsection{User Interface of Data Annotation Platform}

\begin{figure*}[t]
  \centering
  \includegraphics[width=0.95\linewidth]{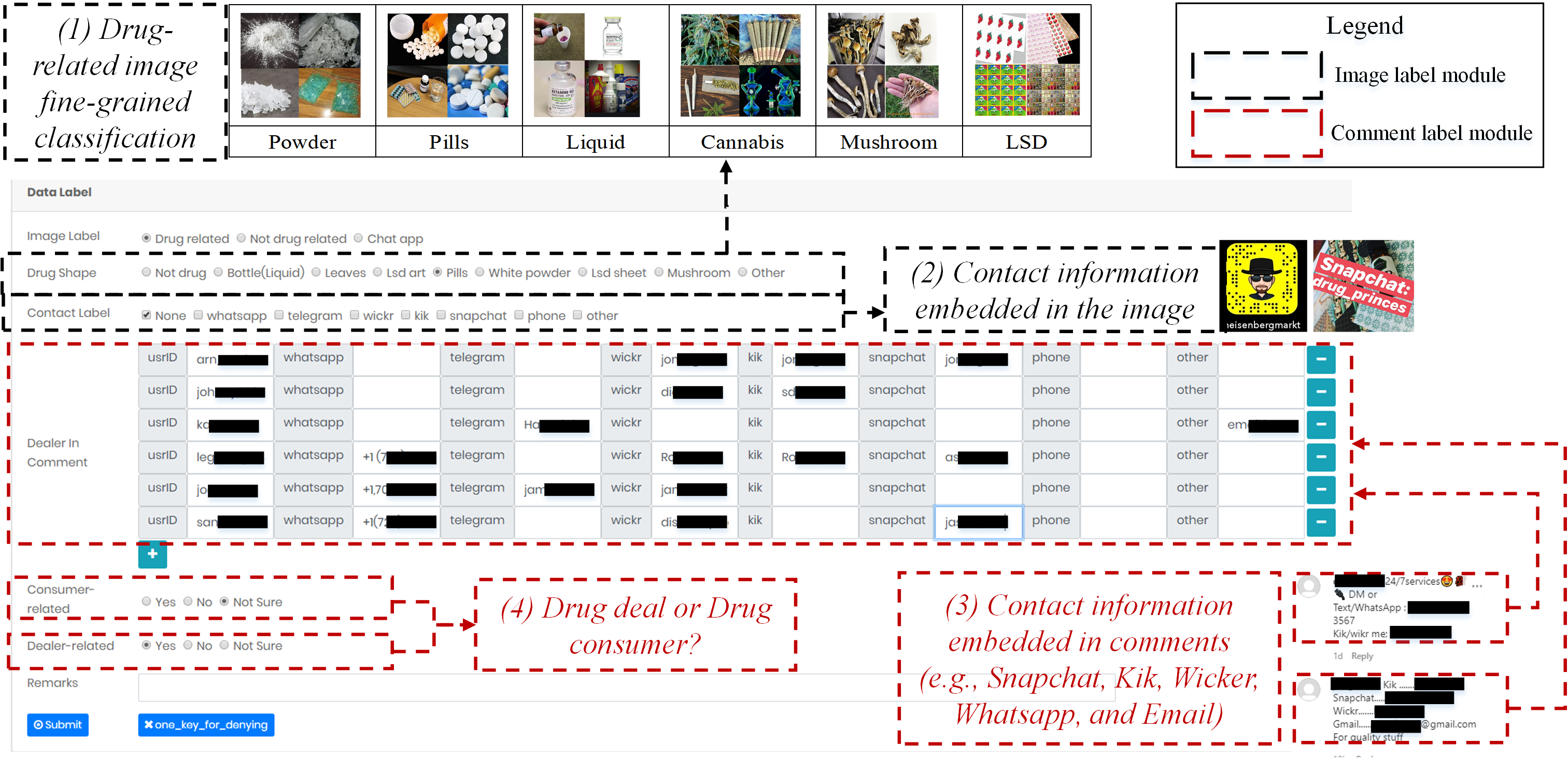}
  \caption{User Interface of Web-based IDDIG Annotation Platform.}
  \Description{}
  \label{fig:3}
\end{figure*}

The collected data can not be used as training data without proper annotation (manual labeling). As the collected posts often contain rich drug trafficking information (e.g., drug types, contact information, sales areas, and so on), we have designed a data annotation platform for drug dealer detection, as shown in Fig. \ref{fig:3}. The annotation platform mainly contains two modules: the image labeling module and comment labeling module. The image labeling module is designed to label image-related information, including fine-grained classification of drug forms (e.g., powder, pills, liquid, cannabis, mushroom, and LSD as shown in Fig. \ref{fig:3}) and contact information (e.g., user ID in the chat app, such as Snapchat, wickr, Kik, What’s up, telegram or email address) embedded in the image. The comment labeling module is designed to label comment-related information, including contact information embedded in the comment, and whether it is related to a drug dealer or a drug consumer. 

\begin{figure*}[t]
  \centering
  \includegraphics[width=\linewidth]{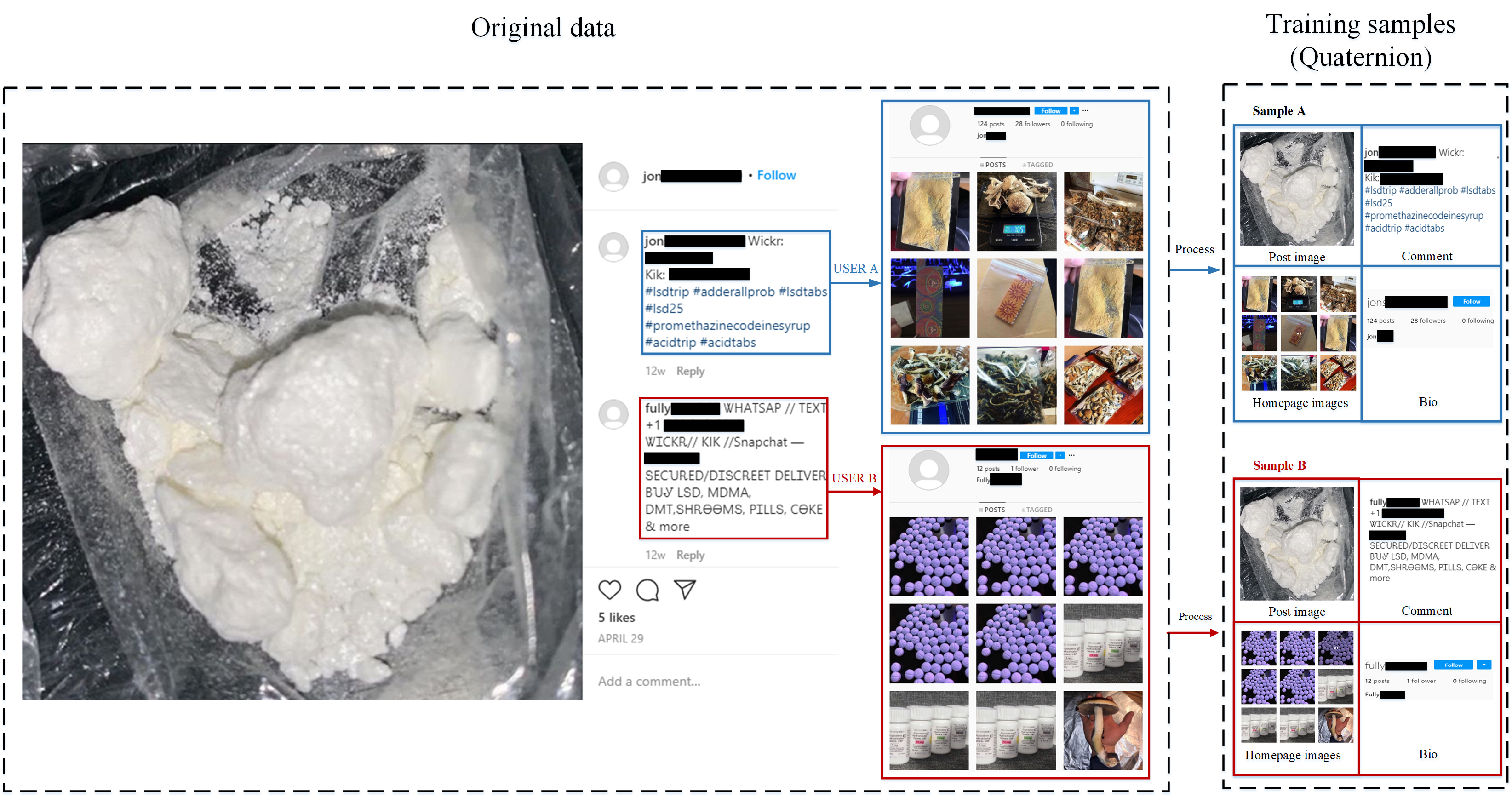}
  \caption{The Conversion Process from Raw Data in IDDIG Dataset to Quadruple-based Training Samples.}
  \Description{}
  \label{fig:7}
\end{figure*}

In addition, for each post, we have also designed a button to check the homepage of users who comment on the post. Based on the post information and homepage information, the domain expert will add an annotation to the post indicating whether it contains a drug dealer account or not. When multiple drug dealer accounts comment on a post, we process the information into multiple data, as shown in Fig. \ref{fig:7}. It takes on average 3-to-5 minutes to label each post; using a crowd-sourcing approach, we have divided the task of data annotation among 10 participants. Overall, we have spent around 400 hours on labeling all collected Instagram data.  

\begin{table}%[b]
  \centering
  \caption{Details of the constructed IDDIG Dataset. }

    \begin{tabular}{clrr}
    \hline
          & Type  & \multicolumn{1}{l}{Number} & \multicolumn{1}{l}{Missing rate} \\
    \hline
    \multirow{6}{*}{Positive} & \# of posts & 1,022 & NA \\
          & \# of unique user accounts & 1,406 & NA \\
          & \# of Posted Image & 2,815 & 0\% \\
          & \# of Posted Comment & 2,722 & 3.30\% \\
          & \# of Homepage Bio & 1,239 & 55.98\% \\
          & \# of Homepage Images & 12,061 & 57.15\% \\
    \hline
    \multirow{6}{*}{Negative} & \# of posts & 1,090 & NA \\
          & \# of unique user accounts & 2,485 & NA \\
          & \# of Posted Image & 3,206 & 0\% \\
          & \# of Posted Comment & 3,206 & 0\% \\
          & \# of Homepage Bio & 2,663 & 16.90\% \\
          & \# of Homepage Images & 24,883 & 22.39\% \\
    \hline
    \end{tabular}%
  \label{tab:tableIDDIG}%
\end{table}%

In summary, a total of 2,112 posts have been annotated and organized for Identifying Drug Dealer  on Instagram (IDDIG).  As shown in Tab. \ref{tab:tableIDDIG}, IDDIG contains 3,206 negative samples and 2,815 positive samples. Each sample was characterized by a quadruple-based data structure, including {\it Posted Comment ($PC$), Posted Image ($PI$), Homepage Bio ($HB$), and Homepage Image ($HI$)}. Among 2,815 positive samples, 1,022 positive samples are pattern 1 (trafficking by direct posting) and the other 1,793 positive samples are pattern 2 (trafficking by indirect commenting) as shown in Fig. \ref{fig:1}. In addition, Tab. \ref{tab:tableIDDIG} shows that more than 50\% positive samples miss homepage bio and homepage images. Most of these missing cases are from pattern 2 samples. To protect the privacy, we manually erase users’ personal information to achieve the purpose of protecting user privacy. Our proposed dataset is publicly available\footnote{Please contact the authors for the dataset}.

\begin{figure}[t]
  \centering
  \includegraphics[width=\linewidth]{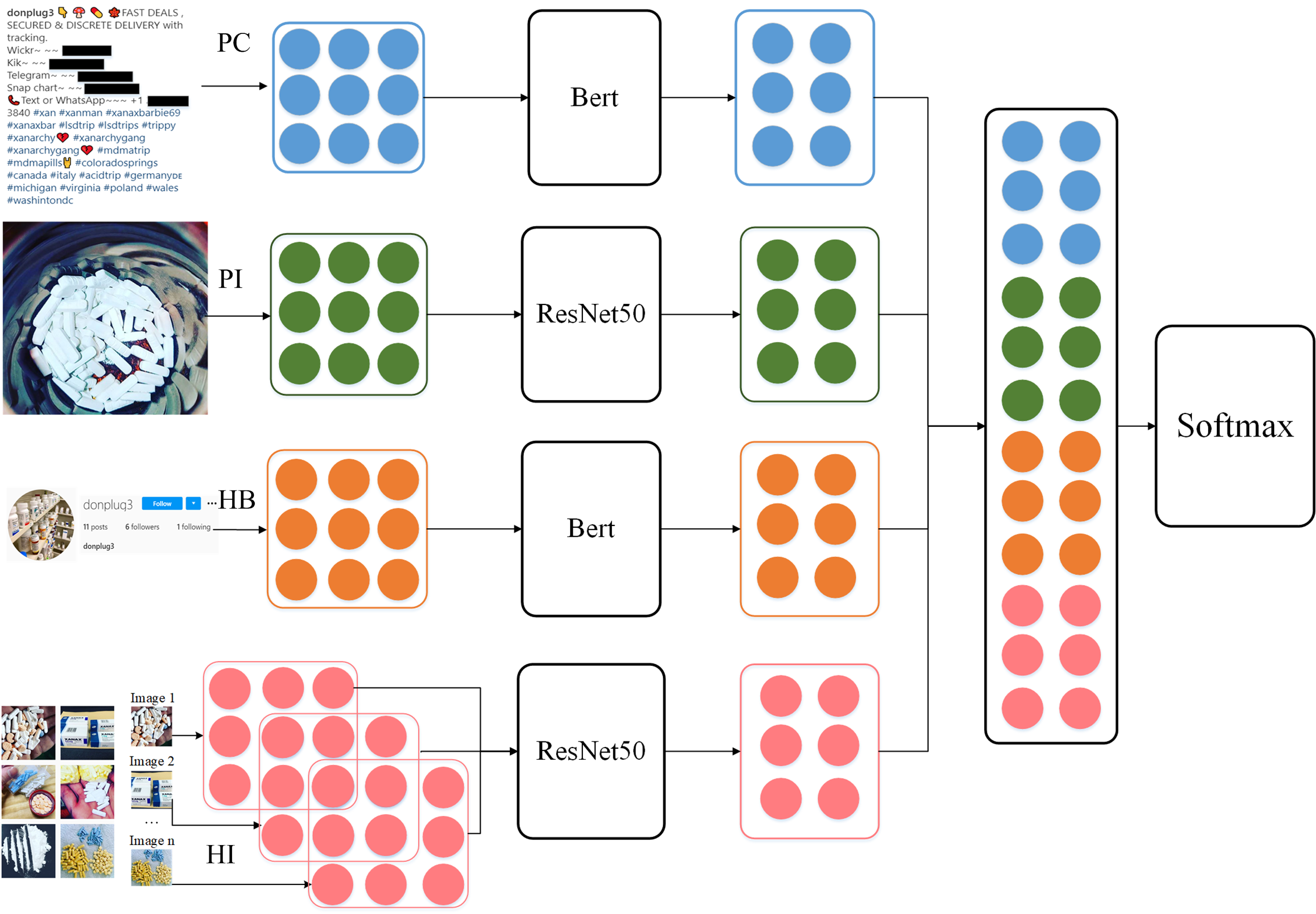}
  \caption{The framework of our proposed multimodal data fusion approach to illicit drug dealer identification. Our model takes in posted comment (PC), posted image (PI), homepage bio (HB), and homepage images (HIs) associated with Instagram users, and combines multimodal information using a quadruple-based fusion strategy to identify an illicit drug dealer.}
  \Description{}
  \label{fig:model}
\end{figure}

%\section{Joint Post and Homepage Identification}
\section{Quadruple-based Multimodal Fusion for Illicit Drug Dealer Identification}
\label{sec:4}

%Furthermore, we extend our research to the application of community detection of drug dealers using unsupervised classification techniques (NetworkX).
\subsection{Problem Formulation and System Overview}
Given the background in Section \ref{sec:3}, we can formally define the problem of \emph{illicit drug dealer identification} as follows: our aim is to build an effective approach to identify illicit drug dealers from normal users on Instagram. For each user, we have her/his posted comment (PC), posted image (PI), homepage bio (HB), and homepage image (HI). Then each user $n$ is represented with quadruple $u^n = \langle PC^n, PI^n, HB^n, HI^n  \rangle$. Let $c^n=1$ denotes that user $N$ is an illicit drug dealer, $c^n=0$ otherwise. Given a set of $n$ training examples $\mathcal D=\{(u^1, c^1), (u^2, c^2), \dots, (u^N, c^N)\}$, we aim at building a predictive model $f: u   \rightarrow  c $ to identify illicit drug dealers. 

We formulate the problem of illicit drug dealer identification as a binary classification problem, namely, the output of our model is the predicted probability $\hat{c}^n$ of a user $n$ being an illicit drug dealer. We have adopted the cross-entropy loss, which has been widely applied in deep learning-based classification \cite{zhang2018generalized}. Specifically, the cross-entropy loss function is defined by
\begin{equation}
    \ell(\Theta) = - \sum_{n=1}^N c^n \log \hat{c}^n.
    \label{eq:1}
\end{equation}

%\textcolor{red}{Definition 4.1 (Illicit drug trafficking activity). An illicit drug trafficking activity (IDTA) is an activity on Instagram that contains the marketing and selling of illicit drugs by posting and commenting a post. It usually contains image and text information. Figure 1 shows some examples of suspect IDTAs on Instagram. Note that a IDTA can be an initial post (e.g., the post initialized by User A). It can also be comments following a post, in which drug trafficking information is added (e.g., comments by Users B). User A usually post image and text to advertise their products and User B leaves some messages on his other chat app ID for drug transaction by commenting. There are no posts on user B’s homepage. We can detect these two IDTAs by combing these features (e.g., drug-related post image, drug-related comment, and drug related homepage.)}
%\label{System Overview.}
%\subsection{System Overview}
Fig. \ref{fig:model} shows the framework of our proposed multimodal data fusion approach to illicit drug dealer identification. Our model takes in posted comments (PC), posted images (PI), homepage bio (HB), and homepage images (HI) associated with Instagram users. A quadruple-based fusion strategy is proposed to  combine the multimodal information, and the combined feature representation is exploited to  identify illicit drug dealers. In this section, we will first discuss the single modality and then elaborate on the strategy of multimodal data fusion.

%Based on the proposed IDDIG dataset from Instagram, we have designed a multimodal fusion method for drug-dealer identification as shown in  Specifically, we first describe the image-based and text-based models, respectively, for drug-dealer identification. Then a multimodal data fusion model is designed to integrate quadruple features to improve the detection performance. 

\subsection{Drug Dealer Identification based on Single Modality}
\subsubsection{Text-based Identification}
Text-based information such as hashtags and captions have been used to differentiate drug-related posts from non-drug-related posts in \cite{yang2017tracking}. It has been shown that there is a clear difference between these two classes of posts; but meanwhile it is easy for Instagram platform administrators to automatically detect these drug-related posts from drug dealer. However, increasingly more drug dealers tend to avoid detect by commenting  under some hot or drug-related posts (see pattern 2 in Fig. \ref{fig:1}). Many cases such cases are collected in our database. As mentioned above, we argue that posted comments and bio information are often more reliable than tags and captions. Therefore, we propose to develop a new text-based classifier based on comments and bio data as follows. To justify the above claim, we have empirically found that comments and biography data posted on Instagram almost always contain many drug-related hashtags, contact information, sales destinations, and payment methods. Such findings are consistent with the common sense that trade-related information, no matter being legal or illicit, has shared characteristics because they are supposed to facilitate the communication between dealers and consumers \cite{li2019machine}. Therefore, we have hand-picked posted comments (including the captions associated with posted images) and bio information at the homepage as the two most salient features in text classifier. Unlike \cite{yang2017tracking}, we propose to leverage the latest advances in natural language processing into the task of text-based drug-dealer identification.

There have been several deep learning-based models with promising performance in text classification, such as CNN \cite{kalchbrenner2014convolutional}, bi-directional long short-term memory (BiLSTM) \cite{zhou2016text}, convolutional long short-term memory (CLSTM) \cite{zhou2015c}, and Bidirectional Encoder Representations from Transformers (BERT) \cite{devlin2018bert}. Among these models, BERT, as a state-of-art text classification model, can consider the full context of a word by looking at the words that come before and after it—particularly useful for understanding the intent behind search queries. BERT uses large-scale corpus to create pretrained language models and fine-tune pretrained models for specific tasks. Therefore, we use the BERT-based model in our method for drug dealer detection. The text information in our IDDIG dataset- namely, comments and bio, are taken as input, respectively, to fine-tune the BERT-based model. The BERT model used  is the BERT-based (12-layer, 768-hidden, 12-heads, 110M parameters). To obtain the sentence-level representation, we extract the token embedding of the last layer and compute the mean vector to get the final feature representation of 768 dimensions.

\subsubsection{Image-based Identification}

\begin{figure}[h]
  \centering
  \includegraphics[width= \linewidth]{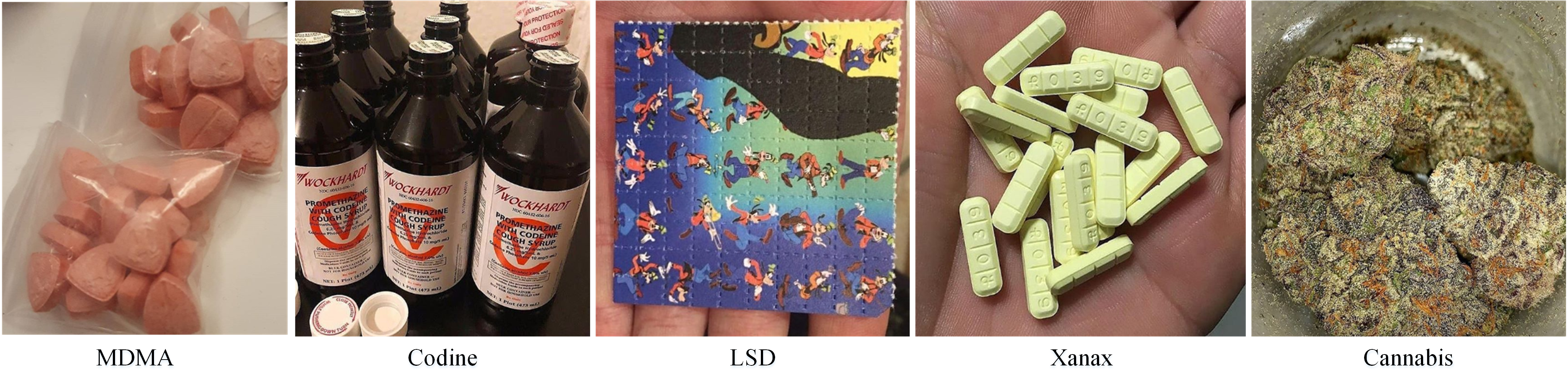}
  \caption{Sample Post Images from IDDIG Dataset (note that they contain diverse drug-related content, from pills and weeds to syrup and powder).}
  \Description{}
  \label{fig:5}
\end{figure}

In previous work \cite{yang2017tracking}, image-based information has been found more reliable than text-based one for the task of drug-related post recognition. Then, the problem of dealer account detection was solved by extracting features from drug-related posts. Thanks to the construction of IDDIG dataset at a large scale, it is possible to directly exploit image information for the task of drug-dealer identification.
Similar to text data, images including both post images and homepage images often contain rich information, such as drug types (as shown in Fig. \ref{fig:5}) and contact information of drug dealers (e.g., user ID from Wickr, Snapchat and telegram). Unlike text data, both drug dealers and regular consumers could post similar images containing drug-related content. Therefore, it is plausible to conjecture that image-based information is less reliable than text-based one for the task of drug-dealer identification. %Our task is to classify images into drug related images and non-drug related images.

As reported later in Sec. \ref{sec:5}, such conjecture has been confirmed by our experimental results. We have used the widely used ResNet-50 \cite{he2016deep} for the task of image-based user classification. ResNet-50 is a pretrained deep learning model for image classification with excellent generalization performance on various recognition tasks \cite{gao2020revisiting,jia2019database}. Thanks to the identity function introduced to the network, the gradient calculation in back-propagation can flow more effectively, which helps alleviate the notorious vanishing gradient problem. Using the collected IDDIG dataset, we have fine-tuned the pretrained ResNet-50 model on our post images and homepage images, respectively, for drug-dealer detection. The images have been cropped to $224 \times 224 \times 3$, and horizontally flipped randomly for data augmentation. We take these processed images as the input and obtain feature vectors in the latent 2048-dimensional space. To implement transfer learning with fine-tuning \cite{pan2009survey}, we have removed the last predicting layer of the pretrained model and replaced them with task-specific prediction layers (drug-dealer vs. non-drug-dealer).

\subsection{Joint Multi-modal Optimization for Drug-Dealer Identification}

Multimodal data fusion \cite{lahat2015multimodal} refers to a class of techniques combining information from multiple modalities for improved performance. In previous work \cite{yang2017tracking}, a decision-level (late) fusion strategy was adopted because training a joint model for feature-level (early) fusion faced several technical challenges such as weak correlation in social media data and lack of paired training data. However, as shown in \cite{gunatilaka2001feature}, feature-level fusion has some advantages over decision-level fusion, at least in theory. Meanwhile, new multimodal data fusion techniques including feature reduction and concatenation \cite{haghighat2016discriminant}, compact and factorized bilinear pooling \cite{lin2015bilinear,gao2020revisiting,jia20203d} have been developed for the applications of multimodal biometric recognition and visual question answering.   

In this work, we propose to leverage these latest advances in multimodal data fusion and explore their potential of feature-level fusion for drug-dealer identification. To the best of our knowledge, this will be the first systematic study of multimodal data fusion, including both protocol design and fusion strategies for drug-dealer identification. We have structured our study into two parts. In the first part, we will present three protocols - namely, multimodal post-based identification, multimodal homepage-based identification, and joint quadruple-based optimization identification. In the second part, we will discuss three competing strategies for multimodal data fusion - feature concatenation \cite{haghighat2016discriminant}, bilinear pooling \cite{lin2015bilinear}, and factorized bilinear pooling \cite{gao2020revisiting,jia20203d}.

\subsubsection{Fusion Protocols}
\label{protocols}
Drug dealers usually post both text and images to promote drugs for sale, as well as comments related to product advertisements, sales destinations, and payment methods. Meanwhile, texts and images can be posted either as comments or at the homepage of an Instagram user. Such multimodal and multisource distribution of collected data distinguishes our quadruple-based data representation from previous works. %Previous works \cite{zhou2016fine,yang2017tracking} have combined post images and comments by a decision-level fusion strategy to improve the performance of drug-related post recognition. Different from existing methods, we propose to fuse both text and image data at the feature-level for drug-dealer identification. 
As shown in Fig. \ref{fig:7}, quadruple-based data representation allows us to come up with three different fusion protocols, as follows.

{\bf 1. Multimodal data fusion}

This is the scenario that has been considered in previous works \cite{zhou2016fine,yang2017tracking}. In addition to feature-level fusion, our approach differs from \cite{zhou2016fine,yang2017tracking} by exploiting the multimodal information contained in both the post and the homepage of Instagram users \cite{hu2014we,jang2015generation}.
The homepage of drug dealers often contains an even richer set of multimodal data than the post, which could serve as supplementary advertising tools. In previous work \cite{yang2017tracking},  homepage statistical data, including percentages of drug-related posts, temporal patterns, relational information, and evidence of transactions are combined to identify a drug dealer account. Instead of using the statistical data derived from the homepage, we propose to fuse the bio text and the last ten post images in homepages directly for drug dealer account identification. For multimodal data fusion at the post level, we have used Resnet-50 and BERT to extract 2048-dimensional image-based and 768-dimensional text-based feature vectors, respectively. For multimodal data fusion at the homepage level, we first extract a sequence of ten image-based features from the ten homepage images, respectively. Then the average of these ten features is used as the final image-based feature representation. If the number of homepage images is less than 10, the features are averaged according to the number of homepage images. If homepage images are missing, the image matrix will be filled with 0 to participate in the multimodal fusion training. Similar to post-level fusion, the averaged 2048-dimensional image-based feature vector will be concatenated with a 768-dimensional text-based feature vector to generate a composite 2816-dimensional vector or combined with text-based feature vector by the bilinear pooling method. The fused feature vector is finally fed into the softmax classifier with cross-entropy loss function in Eq. \eqref{eq:1} for classifying the drug dealer account. 

{\bf 2. Multi-source data fusion}

The collection of multimodal data from different sources (i.e., post vs. homepage) allows us to conduct another line of research on information fusion. Instead of fusing across modalities, it is possible to fuse the information of the same modality but across different sources. Such multisource data fusion \cite{zhang2010multi} has been studied in the field of remote sensing before; under the framework of drug-dealer identification, we believe that multisource data fusion offers complementary insight to multimodal data fusion. Specifically, such comparative studies could help answer the following questions: 1) which modality is more important - image vs. text? 2) which source is more reliable - posts vs. homepage? We conjecture that texts are more important than images due to the intrinsic ambiguity in pictorial representation; posts are more reliable than homepages because posts are often directly used as the advertising tool by drug dealers. Unlike multimodal data fusion, the feature vectors associated with the same modality but from different sources always have the same dimensionality. It is often more convenient to implement FFT-based compact bilinear pooling \cite{fukui2016multimodal} for a pair of equal-length feature vectors. Note that either multimodal or multisource data fusion only reflects one perspective that can be biased; it is desirable to jointly exploit them, as we will elaborate next.

{\bf 3. Joint quadruple-based fusion}

In the real world, drug dealers often need to advertise for drugs while evading being caught by law enforcement. Accordingly, the information we have collected about drug dealers from Instagram is often incomplete - i.e., not all quadruple elements in the proposed IDDIG dataset are available for the same account. Such missing data (a.k.a. missing modality) constraint makes it difficult to use only two-element posts or homepages for accurately identifying drug dealers. The method in \cite{yang2017tracking} first fused multimodal post data to filter drug-related posts, and then detected drug dealer accounts based on multimodal homepage fusion. However, this method often fails to distinguish illicit drug dealers from legal regular consumers or drug abuser accounts. 

Multimodal data fusion with missing data has been recently studied for deep multimodal encoding in \cite{liu2017heterogeneous}. Inspired by the imputation strategy, in this work, we have designed a joint optimization method for quadruple-based information fusion, as shown in Fig. \ref{fig:model}. The features of post-image, comments, bio, and homepage images are first extracted using BERT and ResNet-50 models, respectively. Then the four elementary features are either concatenated into a 5632-dimensional feature vector (the missing element will correspond to an imputation of all zero subvectors) or combined by bilinear pooling operations (note that bilinear pooling can naturally tolerate one missing element and output the available element as the output vector). This way, our quadruple-based fusion can tolerate nine out of a total of 16 different missing patterns. The concatenated or fused features are finally fed into the Softmax for drug dealer detection.

\subsubsection{Fusion Strategies}
\label{strategies}

The strategies of multimodal fusion have been extensively studied in the literature (e.g., \cite{lahat2015multimodal,atrey2010multimodal}). Fusion of textual and visual information has been particularly the focus of research on visual question answering \cite{fukui2016multimodal} and sentiment analysis \cite{chen2017visual}. When feature vectors associated with textual and visual information have different dimensions, concatenating them becomes the easiest solution, even though such strategy suffers from the potential ``curse of dimensionality'' \cite{poggio2017and}. Our experience suggests that such an ad-hoc strategy fits well with the powerful deep learning framework when computational resources are not scarce. A computationally more efficient approach is not to concatenate but to combine multiple feature vectors (assuming normalized to the same length) into a composite vector without the change of dimensionality. This line of research has led to a flurry of bilinear pooling methods, including compact bilinear pooling \cite{fukui2016multimodal} and factorized bilinear pooling \cite{gao2020revisiting}.

In this study, we have conducted a comprehensive comparative study of four competing fusion strategies: 1) {\bf Feature concatenation} \cite{haghighat2016discriminant,zheng2020feature}. Direct concatenation of two feature vectors into a longer one (e.g., $2048+768=2816$).
2) {\bf Bilinear pooling} \cite{lin2015bilinear}. 
Bilinear pooling was introduced in~\cite{lin2015bilinear} to provide a robust image representation for fine-grained image classification. In bilinear pooling, two feature vectors are fused by an outer product (or Kroneker product for matrices) -i.e., we can take element-wise interactions between a pair of feature vectors into account as follows:
\begin{equation}
\boldsymbol Z=\sum_{(i,j\in \mathbb{S})}\boldsymbol{x_}i \boldsymbol{y_}j^{\top}
\label{eq:2}
\end{equation}
where $\left\{\boldsymbol{x_}i|\boldsymbol{x_}i \in \mathbb{R}^p, i \in \mathbb{S} \right\}$, $\left\{\boldsymbol{y_}j|\boldsymbol{y_}j \in \mathbb{R}^q, j\in \mathbb{S} \right\}$ are two feature vectors, $\mathbb{S}$ is the set of spatial locations (combinations of rows and columns), and $\boldsymbol Z\in \mathbb{R}^ {p\times q} $ is the fused feature descriptor. 
Similar to the kernel expansion, it allows the consideration of all pairwise interactions (e.g., $2048*768=1572864$). It can be seen that the size of bilinear feature descriptor can be large, which makes it computationally infeasible. 
3) {\bf Compact bilinear pooling} \cite{fukui2016multimodal}. In view of the explosion of dimensionality, a more compact solution to bilinear pooling is developed based on Fast Fourier Transform (FFT). It requires the two features have identical lengths (e.g., a 768-dimensional vector needs to be resampled to a 2048-dimensional one).
4) {\bf Factorized bilinear coding} \cite{gao2020revisiting}. This is one of the latest advances in computationally efficient bilinear pooling. To generate more compact representations, it is possible to employ a factorized bilinear coding (FBC) strategy to more efficiently integrate the features from multimodal data. %Our experimental findings (refer to Tab. \ref{tab:5}) seem to suggest that the brute-force strategy of feature concatenation achieves the best overall performance except for the precision metric where bilinear pooling achieves the best result. 

The basic idea behind FBC is to encode the features into sparse representations and to learn a dictionary $B$ with $k$ atoms that can be {\em factorized} into low-rank matrices to capture the sparsity structure of observation data. More specifically, let the dictionary $B=[\boldsymbol{b_}1, \boldsymbol{b_}2, ..., \boldsymbol{b_}k] \in \mathbb{R}^{pq\times k} $, FBC proposes to factorize each dictionary atom $\boldsymbol{b_}l\in \mathbb{R}^{pq}
(1\leq l\leq k)$ into $\boldsymbol{U_}l\boldsymbol{V_}l^{\top}$, where $\boldsymbol{U_}l \in \mathbb{R}^{p \times r}$ and $\boldsymbol{V_}l\in \mathbb{R}^{q \times r}$ are learnable low-rank matrices. This way, the original bilinear feature $\boldsymbol{x_}i \boldsymbol{y_}j^{\top}$ can be reconstructed by
$\sum\limits_{l=1}^{k}{c_{v}^{l}}{{\boldsymbol{U}}_{l}}\boldsymbol{V}_{l}^{\top}$, with $\boldsymbol{c_}v \in \mathbb{R}^k$ being the FBC code, and ${c_{v}^{l}}$ being the $l$-th element of $\boldsymbol{c_}v$ ($1\leq v\leq N$, $N$ is the number of pairs in $\mathbb{S}$). 

The sparsity-based FBC encodes two input features $(\boldsymbol{x_}i, \boldsymbol{y_}j)$ into $\boldsymbol{c_}v$ by solving the following optimization problem: 
\begin{equation}
\underset{{{\boldsymbol{c}}_{v}}}{\mathop{\min }}\,\bigg|\bigg|{{\boldsymbol{x}}_{i}}\boldsymbol{y}_{j}^{\top}-\sum\limits_{l=1}^{k}{c_{v}^{l}}{{\boldsymbol{U}}_{l}}\boldsymbol{V}_{l}^{\top}\bigg|{{\bigg|}^{2}}+\lambda||{{\boldsymbol{c}}_{v}}|{{|}_{1}}
\label{eq:3}
\end{equation}
where $\lambda $ is the Lagrangian multiplier controlling the trade-off between the reconstruction error and the sparsity. The FBC code $\boldsymbol{c}_{v}$ can be obtained by the classical LASSO method~\cite{tibshirani1996regression} which produces
\begin{equation}
\left\{ \begin{array}{l}
  \boldsymbol{{c}'}_{v}=\boldsymbol{P}({{\boldsymbol{\widetilde U}}^{\top}}{\boldsymbol{x}_{i}}\circ{{\boldsymbol{\widetilde V}}^{\top}}{{\boldsymbol{y}}_{j}}), \\ 
 {{\boldsymbol{c}}_v}={\rm sign}({\boldsymbol{c}'_v})\circ {\rm max}(({\rm abs}({\boldsymbol{c}'_{v}})-\frac{\lambda }{2}),0). \\
\end{array} 
\right.
\label{eq:4}
\end{equation}
where $\circ$ denotes the Hadamard product, $\boldsymbol{P}\in \mathbb{R}^{k\times rk}$ is a binary matrix with only elements in the row $l$, columns $((l-1)r+1)$ to $(lr)$ being ``1", and $\boldsymbol{\widetilde U}$ and $\boldsymbol{\widetilde V}$ are the transformations of $\boldsymbol{U}$ and $\boldsymbol{V}$ are in the form of
\begin{equation}
\left\{ \begin{array}{l}
  {\boldsymbol{\widetilde U}^{\top}}=[\boldsymbol{\widetilde U}_{l}^{\top}] =[\frac{1}{r}\cdot{\boldsymbol{I}}(({\boldsymbol{q}_{l}}\boldsymbol 1_{rk}^{\top})\circ{{\boldsymbol U}^{\top}})]\in {{\mathbb{R}}^{rk\times p}}\\
  {\boldsymbol{\widetilde V}^{\top}}=[\boldsymbol{\widetilde V}_{l}^{\top}] =[\frac{1}{r}\cdot {\boldsymbol{I}}{{\boldsymbol{V}}^{\top}}]\in {\mathbb{R}^{rk\times q}}
\end{array} 
\right.
\label{eq:5}
\end{equation}
where $\boldsymbol{I}\in {\mathbb{R}}^{r\times rk}$ is an all ``1" matrix, $\boldsymbol{q}_{l}$ is the $l$-th column of $\boldsymbol Q=((\boldsymbol P(\boldsymbol U^{\top}{\boldsymbol U}\boldsymbol P^{\top}\cdot \boldsymbol V^{\top}{\boldsymbol V}\boldsymbol P^{\top}))^{-1}{\boldsymbol P})^{\top}$. 
%With Eq.~\eqref{eq:3}, the FBC code $\boldsymbol{c}_{v}$ can be obtained by learning $\boldsymbol{\widetilde U}$ and $\boldsymbol{\widetilde V}$ instead of $\boldsymbol{U}$ and $\boldsymbol{V}$. We can get all FBC codes $\boldsymbol{c} $ in feature pair set $\mathbb{S}$; then they are fused using the max operation to attain the final global representation $z=max\left\{\boldsymbol{c}_{v}\right\}_{i=1}^{N}$.

%\subsection{Illicit Drug Dealer Classification}

\section{Unsupervised Learning for Drug-Dealer Community Detection}
\label{sec:4a}

Community detection in a social network refers to the problem of identifying a cluster of connected vertices forming fairly independent compartments in a graph or a network \cite{newman2006modularity,newman2004finding}. Detecting drug-dealer communities (a.k.a. drug trafficking rings) in the real world has been studied by social scientists and criminologists for decades \cite{adler1993wheeling}. Finding the community of drug dealers and their customers from social media data such as Instagram has only been recently considered (e.g., \cite{li2019machine}). By exploring how the hashtags are connected to Instagram, we make the first step toward revealing the hashtag culture, which might lead to useful hints for online community detection. In other words, the complicated relational characteristics of Instagram hashtag can be discovered through data mining and visualization techniques, allowing a deeper understanding of drug-dealer communities/clusters found across hashtags.

To demonstrate the feasibility of this idea, we have designed an unsupervised learning method using NetworkX \cite{hagberg2008exploring} - a graph object and a Python language package developed for exploiting and analyzing networks. NetworkX has been widely used for operations on large-scale real-world graphs, such as graphs with over 10 million nodes and 100 million edges \cite{hagberg2010hacking}. For the IDDIG dataset, we have collected from Instagram, more than 800 posts in our dataset all contain at least one mention of drug-related hashtag (e.g., \#pills, \#psychedelic, \#mdmatherapy, \#mdmatrip, \#lsdtrip, and \#lsdls). All unique hashtags in the corpus represent nodes in the graph. Edges were formed between hashtags if they were mentioned together in the same post. This yielded an undirected drug-related graph, which was then analyzed by NetworkX. For the task of community detection, we can calculate various graph attributes such as betweenness centrality \cite{brandes2001faster}, adjacency matrix \cite{papadopoulos2012community}, clustering coefficient \cite{lancichinetti2009community}, and vertex similarity \cite{fortunato2010community}. The identified communities in a network are often visualized by a tool, so-called Sunburst plot (please refer to Fig. \ref{fig:8} for illustrative examples).

\section{Experimental Results}
\label{sec:5}

In this section, we first report the performance of single modality methods on our own IDDIG dataset for drug-dealer identification. Different multimodal fusion schemes (protocols and strategies) were then compared and analyzed. Finally, we present our experimental results of drug-dealer community detection using NetworkX and analyzed the characteristics of drug-dealer community on Instagram.

\subsection{Experimental Setup}
The IDDIG dataset used in our experiments contains 3,206 negative samples and 2,815 positive samples. As shown in Tab. \ref{tab:my-table}, the size of positive user accounts is much larger than all existing datasets. We have split the whole IDDIG dataset into 70\% training set and 30\% testing set. For post and homepage images, we have adopted a ResNet50 to extract the feature vectors of length 2048; for posted comments and bio texts, we have used BERT pretrained model to extract a 768-dimensional vector. We have trained the proposed multimodal fusion model using the popular Adam optimization algorithm \cite{reddi2019convergence} with a mini-batch size of 10 based on IDDIG. The following parameters are adopted in our setting: learning rate $\alpha=0.001$, $\beta_1=0.9,\beta_2=0.999$, $\epsilon=10^{-8}$. We opt to terminate the training after 50 epochs. All experiments are conducted using PyTorch on a workstation with one RTX 2080 GPU.

%data source based, data type based, and quaternion based fusion scheme. 
%(1) Data source based fusion scheme: As the quaternion data source from post and homepage, we are committed to fusing the post and homepage data separately to compare which information is more important for the drug dealer detection. In this fusion scheme, we designed post-based fusion sub-scheme and homepage-based fusion. 

%(2) Data type based fusion scheme: To prove which feature of image and text has more influence on drug detection, we designed image-based fusion subscheme and text-based fusion subscheme.

%(3) Quaternion based fusion scheme: We designed early fusion and late fusion based on quaternion data, and compare their classification effects in drug detection. Early fusion: 

\subsection{Drug-dealer Identification Results}

%\subsubsection{Experimental Setup}
\begin{table}[]
\caption{Text-based Classification for Drug Dealer Detection}
\label{tab:tab1_CommentClassification}
\begin{tabular}{llllll}
\hline
Performance & Accuracy  & Precision & Recall  & F1 score   \\
\hline
PC only        &        &        &        &        \\
CNN\cite{kim2014convolutional}    & 84.38\% & 69.12\% & 95.05\% & 80.03\% \\
BiLSTM\cite{chen2017improving}    & 83.42\% & \textbf{94.00\%} & 75.45\% & 83.71\% \\
CLSTM\cite{zhou2015c}             & 85.12\% & 70.16\% & \textbf{95.90\%} & 81.04\% \\
BERT\cite{devlin2018bert}         & \textbf{87.17\%} & 93.67\% & 75.98\% & \textbf{83.90\%} \\
\hline
HB only         &        &        &        &        \\
CNN\cite{kim2014convolutional}    & 84.51\% & 67.32\% & \textbf{97.82\%} & 79.75\% \\
BiLSTM\cite{chen2017improving}    & 84.04\% & 73.46\% & 89.42\% & 80.66\% \\
CLSTM\cite{zhou2015c}             & \textbf{84.51\%} & \textbf{93.47\%} & 70.76\% & 80.55\% \\
BERT\cite{devlin2018bert}         & 84.38\% & 76.52\% & 93.05\% & \textbf{83.98\%} \\
\hline
\end{tabular}
\label{tab:2}
\end{table}

\subsubsection{Single modality based Results for Drug-Dealer Detection}
To evaluate the performance of single modality classification at this stage by training and testing on the proposed IDDIG dataset. We have implemented four models (i.e., CNN\cite{kim2014convolutional}, BiLSTM\cite{chen2017improving}, CLSTM \cite{zhou2015c} and BERT\cite{devlin2018bert}) for text classification and compare their performance based on the proposed IDDIG dataset. Four different performance metrics (accuracy, precision, recall, and F1 score) are reported in Tab. \ref{tab:2}. It can be observed that 1) BERT has the best performance than any other text classification model with 83.90\% F1 score on comment and 83.98\% F1 score on bio; this is not surprising because BERT represents the current state-of-the-art in natural language processing. 2) comment-based and bio-based textural information have similar discrimination power because they have both achieved over 80\% F1 score. 3) two LSTM models (BiLSTM\cite{chen2017improving} and CLSTM \cite{zhou2015c}) have demonstrated strikingly different precision/recall results, which reflect the trade-off between these two objectives.

\begin{table}[]
\caption{Image-based Classification for Drug Dealer Detection}
\label{tab:imageResult}
\begin{tabular}{lllll}
\hline
Performance             & Accuracy & Precision & Recall  & F1 score \\
\hline
PI only     &          &           &         &          \\
VGG16 \cite{simonyan2014very}       & 70.72\%  & 63.44\%   & 70.94\% & 66.98\%  \\
ResNet50 \cite{he2016deep}          & 69.69\%  & 62.61\%   & 68.47\% & 65.41\%  \\
ResNet152 \cite{he2016deep}         & \textbf{73.61\%}  & 65.96\%  & \textbf{76.35\%} & \textbf{73.61\%}  \\
DenseNet121 \cite{huang2017densely} & 72.78\%  & \textbf{67.49\%}   & 67.49\% & 67.49\%  \\
\hline
HI only  &          &           &         &          \\
VGG16 \cite{simonyan2014very}        & 83.25\%  & 74.94\%   & 93.05\% & 83.02\%  \\
ResNet50 \cite{he2016deep}           & \textbf{84.04\%}  & \textbf{76.11\%}   & 92.90\% & \textbf{83.67\%}  \\
ResNet152 \cite{he2016deep}          & 83.58\%  & 75.09\%  & \textbf{93.80\%} & 83.41\%  \\
DenseNet121 \cite{huang2017densely}  & 83.37\%  & 75.93\%  & 93.61\%   &83.27\%      \\
\hline
\end{tabular}
\label{tab:3}
\end{table}

For image-based classification, we have trained four models(i.e., VGG-16 \cite{simonyan2014very}, ResNet-50 \cite{he2016deep}, ResNet-152 \cite{he2016deep}, and DenseNet121 \cite{huang2017densely}) to classify our post images and homepage images. As shown in Tab. \ref{tab:3}, the best performance for post image classification is achieved by ResNet-152 with over 73\% accuracy and 73\% F1 score and the best performance for homepage image classification is achieved by ResNet-50 with over 84\% accuracy and 83\% F1 score. When compared with homepage images, we observe that the performance of image-based classification is noticeably lower for post images. Such an experimental finding is consistent with our common sense because homepage is often a more reliable source than post for drug-dealer identification (note that we have used ten images in the homepage). In addition, our experimental results show that homepage image-based classifications have achieved comparable performance to text-based (closer to bio-based than post-based). Previous study \cite{yang2017tracking} has demonstrated that accounts with a higher value of drug-related percentage are more likely to be drug dealers. Our study further corroborates such findings thanks to numerous drug-related images posted at the homepage, which justifies the benefit of our data collection from the homepage.

\subsubsection{Fusion based Results for Drug-Dealer Detection}
To compare and analyze different multimodal data fusion methods, we have designed three fusion experimental schemes including data-source-based fusion, data-type-based fusion, and quadruple-based fusion. We will first report our experimental results in correspondence with the three protocols discussed in Sec. \ref{protocols}. Then, we will compare different fusion strategies, as covered in Sec. \ref{strategies}. 

% Fusion experimental result 
\begin{table*}[t]
  \newcommand{\tabincell}[2]{\begin{tabular}{@{}#1@{}}#2\end{tabular}} 
  \centering
  \caption{Comparison of Different Fusion Protocols (top-down): Multi-modal Data Fusion, Multi-source Data Fusion, and Quadruple-based Fusion.}
    \begin{tabular}{llrrrr}
    \hline
    \multicolumn{2}{c}{Performance} & \multicolumn{1}{l}{Accuracy} & \multicolumn{1}{l}{Precision} & \multicolumn{1}{l}{Recall} & \multicolumn{1}{l}{F1 score} \\
    \hline
    Post-level fusion & PI + PC & 94.95\% & 93.47\% & 95.17\% & 94.31\% \\
    Homepage-level fusion & HI + HB & 91.29\% & 88.20\% & 92.60\% & 90.35\% \\
    \hline
    Text-based fusion & PC + HB   & 93.68\% & 90.44\% & 95.77\% & 93.03\% \\
    Image-based fusion & PI + HI & 92.95\% & 87.57\% & \textbf{97.89\%} & 92.44\% \\
    \hline
    \tabincell{l}{Quadruple-based fusion \\ (Decision-level fusion)} & \tabincell{l}{PI + PC + \\ HI + HB} & 94.48\% & \textbf{96.92\%} & 90.33\% & 93.51\% \\
    \tabincell{l}{Quadruple-based fusion \\ (Feature-level fusion)} & \tabincell{l}{PI + PC + \\ HI + HB} & \textbf{95.55\%} & 96.41\% & 93.35\% & \textbf{94.86\%} \\
    \hline
    \end{tabular}%
  \label{tab:4}%
\end{table*}%

(1) Multi-modal data fusion: In a previous study \cite{yang2017tracking}, it has been shown that the fusion of image and text data can improve the performance of drug-related post recognition. Since we have collected multimodal data from different sources (post vs. homepage), we have conducted the fusion experiments at both post-level and homepage level. The experiment results are as shown at the top row in Table \ref{tab:4} have confirmed the benefit of multimodal data fusion in terms of significant performance improvement. Moreover, our results have demonstrated that post-based fusion may have the better performance with over 94\% accuracy and 94\% F1 Score than homepage-based fusion. One possible explanation for such findings is that textual information plays a more dominating role in drug-dealer identification than visual information. For example,  contact information along such as user ID is often sufficient because only drug dealers prefer financial incentives to privacy concerns.

% Compare post multimol fusion with bilinear pooling
\begin{table}[htbp]
  \centering
   \newcommand{\tabincell}[2]{\begin{tabular}{@{}#1@{}}#2\end{tabular}}
  \caption{Comparison of Experimental Results between Feature Concatenation and Various Bilinear Pooling Methods.}
    \begin{tabular}{llllll}
    \hline
    \multicolumn{1}{c}{Performance} & \multicolumn{1}{l}{Accuracy} & \multicolumn{1}{l}{Precision} & \multicolumn{1}{l}{Recall} & \multicolumn{1}{l}{F1 score} &  \\
    \hline
    Bilinear Pooling & 94.02\% & \textbf{95.40\%} & 90.79\% & 93.03\% &  \\
    \hline
   \tabincell{l}{Compact \\Bilinear Pooling} & 93.62\% & 92.88\% & 92.60\% & 92.74\% &  \\
    \hline
    \tabincell{l}{Factorized \\Bilinear Pooling} & 92.02\% & 93.85\% & 87.61\% & 90.63\% &  \\
    \hline
    Concat (ours) & \textbf{94.95\%} & 93.47\% & \textbf{95.17\%} & \textbf{94.31\%} &  \\
    \hline
    \end{tabular}%
  \label{tab:5}%
\end{table}%

(2) Multi-source data fusion: An alternative fusion protocol is across different sources instead of modalities. We have implemented text-based feature fusion with 768 + 768 dimensions by concatenating two BERT model output vectors based on the comment and bio in IDDIG dataset. We have also implemented image-based feature fusion with 2048 + 2048 dimensions by concatenating two ResNet model output vectors based on the post-image and homepage images. It can be seen that text-based feature fusions produce better classification results with over 93\% accuracy and 93\% F1 score than image-based feature fusion, as shown in the middle row in Tab. \ref{tab:4}. It is interesting to note that the recall performance is notably better than the precision performance for image-based fusion. Such low-precision and high-recall observation is consistent with the nature of visual information - i.e., it is too liberal for the task of identifying drug dealers.  By contrast, textual information is more conservative and favorable to the precision metric as adopted by a related work \cite{li2019machine}. This is because text data on Instagram do contain more direct drug-trafficking related information, including various uses of drug-related hashtags, evidence of transactions and payment methods. %This makes it more conducive to model learning and identification of drug trafficking characteristics.

(3) Quadruple-based fusion: Similar to the previous work \cite{gunatilaka2001feature}, we have implemented and compared feature-level fusion (ours) with decision-level fusion (adopted by \cite{yang2017tracking}) on the IDDIG dataset. In our implementation, a linear weighting method is used for decision-level fusion (similar to \cite{yang2017tracking}). Without any prior knowledge, we have selected 25\% weights for post-image--image-based classifier, comment-based classifier, bio-based classifier, and homepage image-based classifier, respectively. The performance comparison result of the two fusion methods (decision-level vs. feature-level) is shown at the bottom row in Table \ref{tab:4}. It demonstrated that the proposed feature-level fusions in this paper have better accuracy and F1 score results than decision-level fusion.  Feature-level fusions have some advantages over decision-level fusion strategies, in that the process of decision-making cannot increase the amount of information contained in the input features (due to data processing inequality \cite{cover1999elements}).

(4) Different feature-level fusion strategies: 
As discussed in Sec. \ref{strategies}, information fusion at the feature level can be implemented by several strategies. As of today, there still lacks a rigorous theory for evaluating and comparing these competing strategies. Therefore, we have opted to take an empirical approach in this study. Four competing fusion strategies have been implemented: feature concatenation \cite{haghighat2016discriminant}, bilinear pooling \cite{lin2015bilinear}, compact bilinear pooling \cite{fukui2016multimodal}, and factorized bilinear pooling\cite{gao2020revisiting}.
As shown in Tab. \ref{tab:5}, feature concatenation has achieved the overall best performance (only falls behind bilinear pooling on the precision metric). Such findings suggest that brute-force concatenation is capable of outperforming more delicate pooling as long as the increase of dimensionality remains manageable.

\begin{figure}[t]
  \centering
  \includegraphics[width=0.6 \linewidth]{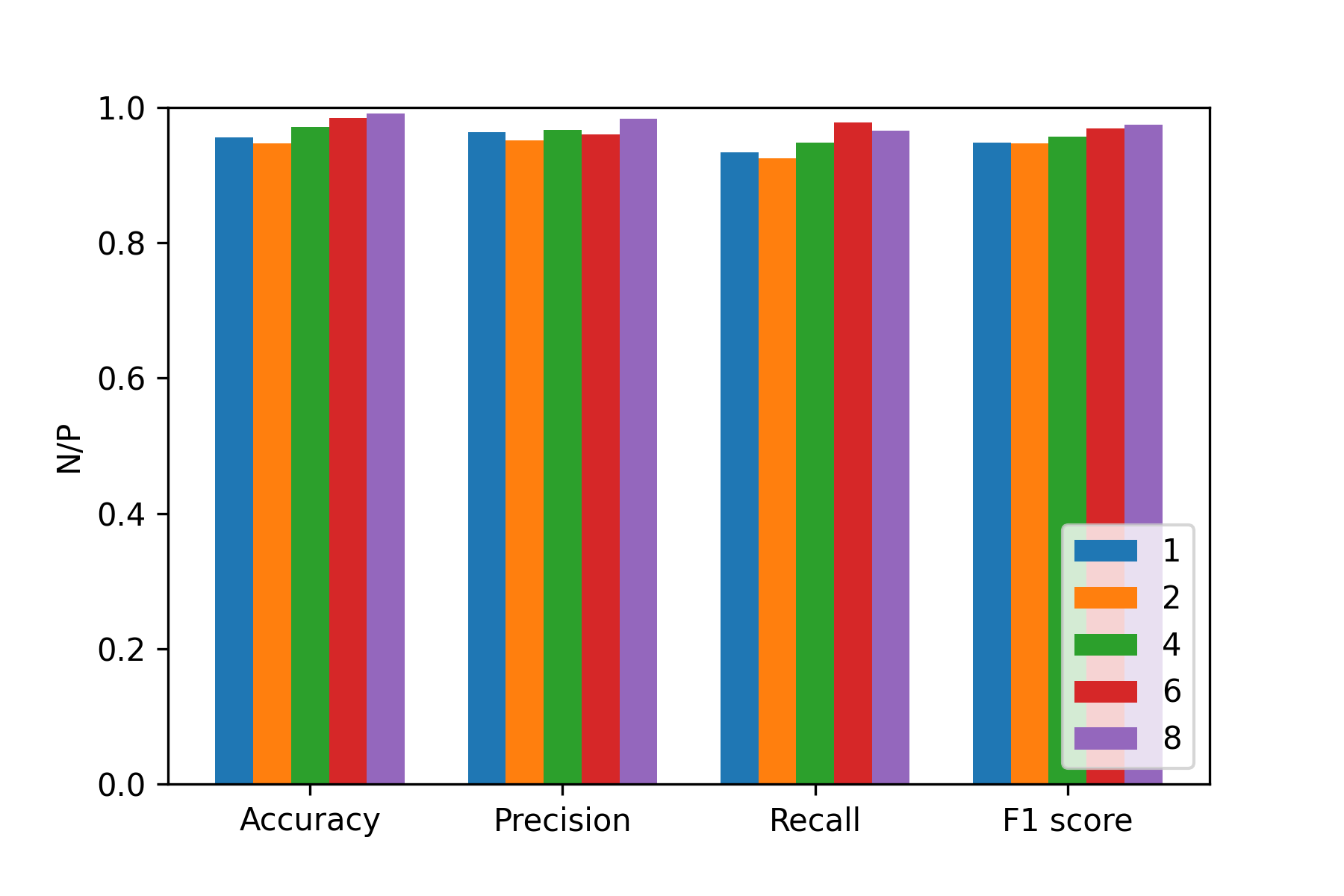}
  \caption{Impacts of Negative/Positive Ratio on performances of illicit drug dealer identification.}
  \Description{}
  \label{fig:ratio_impact}
\end{figure}

\subsection{Impact of negative-positive ratio}
In the real world scenario, the portion of drug dealers on Instagram is small, which means the dataset can be highly imbalanced. To make our dataset closer to the real world distribution (the number of negative samples is larger than positive), we have collected many nondrug dealers and their associated data on Instagram. Such data augmentation allows us to study the impact of negative-positive ratio on the identification performance of our system. Specifically, we have added different number of negative samples to compare and evaluate the robustness of quadruple-based fusion method. The ratio of negative data over positive data ($N/P$) is set to 2$:$1, 4$:$1, 6$:$1 and 8$:$1, and the identification results are shown in Fig. \ref{fig:ratio_impact}. 
We found our model is not sensitive to negative-positive ratios in the testing range (note that a much larger ratio than 8 might better characterize the challenge of ``finding a needle in a haystack'' in practice).
%We can find that model obtained better performance (higher F1) with more negative samples. It demonstrates that model performance is still good based on unbalance training dataset. The comparison of confusion of different $N/P$ training dataset shows that the proposed quadruple-based fusion has a good robustness (See in Fig. \ref{fig:8}).} 

%\vspace{-0.2in}
\subsection{Drug-related community detection and visualization}

\begin{figure*}%[th]
  \centering
  \includegraphics[width=0.3\linewidth]{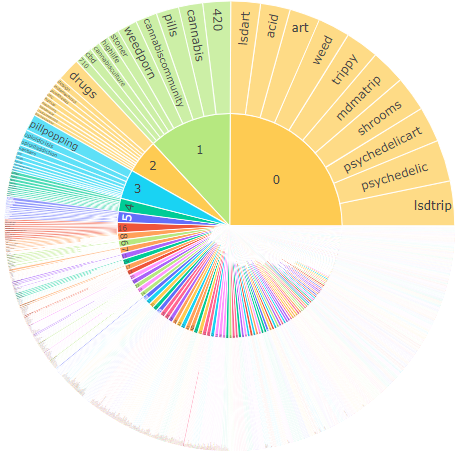}
  \includegraphics[width=0.59\linewidth]{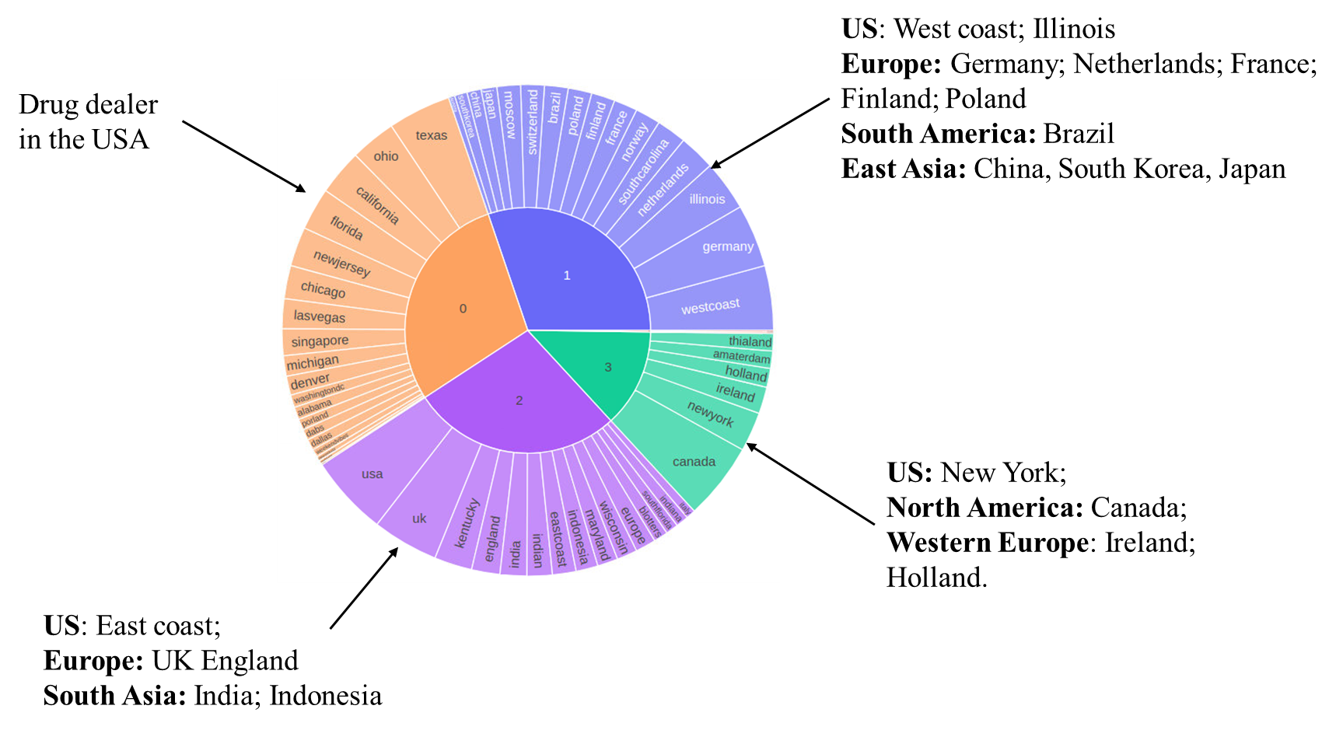}
  
  a) \hspace{6.5cm} b)
  \caption{Sunburst Plots of Community Detection from IDDIG Dataset. a) Network Structure Discovered by Drug-type Hashtag; b) Community Structure Detected by Geography Information Using Hashtag Xanax.}
  \Description{}
  \label{fig:8}
\end{figure*}

The constructed IDDIG dataset also allows us to discover drug-related community structures, which might shed novel insights on the dynamics and evolution of illicit drug trade in the real world. In our experiment, we have first conducted a hashtag-based community detection algorithm using the existing tool NetworkX \cite{hagberg2008exploring}. The default parameter setting has been used (e.g., at most 10 most frequent nodes in each cluster will be preserved). Fig. \ref{fig:8} a) shows the sunburst plot of IDDIG dataset organized by different drug types. It can be observed that the class of Lysergic acid diethylamide (LSD) represents the most popular community on Instagram (accounting for almost 25\% of hashtags). This is a novel discovery, supporting the strong correlation with the young age of Instagram users, such as teenagers. Based on the hashtag ``\#Xanax'', we have further analyzed the community structure based on the available geography information (i.e., place-related hashtags). Fig. \ref{fig:8} b) shows the sunburst plot of Xanax hot spots discovered from IDDIG data. We note that the states of Texas, Ohio, and California are the top three ranked in terms of drug-dealer locations. In the meantime, Instagram data also reveals worldwide connections of illicit drug trade between US and other countries in Europe, South America, and Asia. 

\section{CONCLUSION}
\label{sec:6}

\noindent \emph{Concluding Remarks.} In this study, we have collected and constructed a large-scale dataset (IDDIG) from Instagram data to support the research related to drug-dealer identification. Our dataset includes both textual and visual information contained in posted comments as well as at the homepage. An automatic hashtag-based iterative data crawling system and a user-friendly interactive web-based data annotation system were presented. Our data crawling and annotation systems allow us to build a dataset with thousands of positive and negative samples. Based on the constructed IDDIG dataset, we have developed a quadruple-based data representation for drug-dealer identification. BERT and ResNet models are adopted to extract text-based and image-based features from quadruple-based data representation, respectively. We have conducted a comprehensive study of multimodal data fusion for the identification of drug dealers and verified the performance improvement due to quadruple-based fusion. The overall accuracy of our drug-dealer identification has reached almost 95\% on the IDDIG dataset, according to our experimental results.  Furthermore, we have detected the drug dealer account community by mining hashtag culture. The experiments show that our proposed approach is effective and able to detect drug dealer accounts with high accuracy thanks to data fusion across modalities and sources. 

\noindent

\emph{Limitations and Future Research.} There are a few
limitations in this work and interesting future research
directions. First, missing modalities are common in the dataset. We adopted an ad-hoc imputation strategy for the missing modality problem in this work. An interesting future work is to explore more efficient fusion strategies with incomplete modalities. Generative adversarial network (GAN)-based approaches have shown promising performance on handling missing modalities in \cite{shang2017vigan}. However, how to address this issue under the framework of multimodal data fusion has remained open \cite{ma2021smil}. Second, although we have assessed the impact of negative-positive ratio on  performances of illicit drug dealer identification, in real practice, the negative-positive ratio can be much more negatively skewed. The model can be better  evaluated with negative-positive ratio that approximates the real ratio on Instagram. Third, robust detection of drug-related communities such as drug ring especially how to establish the connection between darkweb and clearweb is a challenging problem, which will be left for future studies.

\begin{acks}
This work is partially supported by the NSF under grants IIS-2107172, IIS-2027127, IIS-2040144, CNS-2034470, IIS-1951504, CNS-1940859, CNS-1814825 and OAC-1940855, and the DoJ/NIJ under grant NIJ 2018-75-CX-0032. 
\end{acks}

\bibliographystyle{ACM-Reference-Format}
%\bibliography{ACM_TIST.bib}
%\bibliographystyle{acm}
\bibliography{ref}


\begin{thebibliography}{86}

%%% the Web address.

\ifx \showCODEN    \undefined \def \showCODEN     #1{\unskip}     \fi
\ifx \showDOI      \undefined \def \showDOI       #1{#1}\fi
\ifx \showISBNx    \undefined \def \showISBNx     #1{\unskip}     \fi
\ifx \showISBNxiii \undefined \def \showISBNxiii  #1{\unskip}     \fi
\ifx \showISSN     \undefined \def \showISSN      #1{\unskip}     \fi
\ifx \showLCCN     \undefined \def \showLCCN      #1{\unskip}     \fi
\ifx \shownote     \undefined \def \shownote      #1{#1}          \fi
\ifx \showarticletitle \undefined \def \showarticletitle #1{#1}   \fi
\ifx \showURL      \undefined \def \showURL       {\relax}        \fi
% invisible to TeX
\providecommand\bibfield[2]{#2}
\providecommand\bibinfo[2]{#2}
\providecommand\natexlab[1]{#1}
\providecommand\showeprint[2][]{arXiv:#2}

\bibitem[\protect\citeauthoryear{Adler}{Adler}{1993}]%
        {adler1993wheeling}
\bibfield{author}{\bibinfo{person}{Patricia~A Adler}.}
  \bibinfo{year}{1993}\natexlab{}.
\newblock \bibinfo{booktitle}{\emph{Wheeling and dealing: An ethnography of an
  upper-level drug dealing and smuggling community}}.
\newblock \bibinfo{publisher}{Columbia University Press}.
\newblock


\bibitem[\protect\citeauthoryear{Atrey, Hossain, El~Saddik, and
  Kankanhalli}{Atrey et~al\mbox{.}}{2010}]%
        {atrey2010multimodal}
\bibfield{author}{\bibinfo{person}{Pradeep~K Atrey}, \bibinfo{person}{M~Anwar
  Hossain}, \bibinfo{person}{Abdulmotaleb El~Saddik}, {and}
  \bibinfo{person}{Mohan~S Kankanhalli}.} \bibinfo{year}{2010}\natexlab{}.
\newblock \showarticletitle{Multimodal fusion for multimedia analysis: a
  survey}.
\newblock \bibinfo{journal}{\emph{Multimedia systems}} \bibinfo{volume}{16},
  \bibinfo{number}{6} (\bibinfo{year}{2010}), \bibinfo{pages}{345--379}.
\newblock


\bibitem[\protect\citeauthoryear{Barbier and Liu}{Barbier and Liu}{2011}]%
        {barbier2011data}
\bibfield{author}{\bibinfo{person}{Geoffrey Barbier} {and}
  \bibinfo{person}{Huan Liu}.} \bibinfo{year}{2011}\natexlab{}.
\newblock \showarticletitle{Data mining in social media}.
\newblock In \bibinfo{booktitle}{\emph{Social network data analytics}}.
  \bibinfo{publisher}{Springer}, \bibinfo{pages}{327--352}.
\newblock


\bibitem[\protect\citeauthoryear{Brandes}{Brandes}{2001}]%
        {brandes2001faster}
\bibfield{author}{\bibinfo{person}{Ulrik Brandes}.}
  \bibinfo{year}{2001}\natexlab{}.
\newblock \showarticletitle{A faster algorithm for betweenness centrality}.
\newblock \bibinfo{journal}{\emph{Journal of mathematical sociology}}
  \bibinfo{volume}{25}, \bibinfo{number}{2} (\bibinfo{year}{2001}),
  \bibinfo{pages}{163--177}.
\newblock


\bibitem[\protect\citeauthoryear{Buntain and Golbeck}{Buntain and
  Golbeck}{2015}]%
        {buntain2015your}
\bibfield{author}{\bibinfo{person}{Cody Buntain} {and}
  \bibinfo{person}{Jennifer Golbeck}.} \bibinfo{year}{2015}\natexlab{}.
\newblock \showarticletitle{This is your Twitter on drugs: Any questions?}. In
  \bibinfo{booktitle}{\emph{Proceedings of the 24th international conference on
  World Wide Web}}. \bibinfo{pages}{777--782}.
\newblock


\bibitem[\protect\citeauthoryear{Chen, Xu, He, and Wang}{Chen
  et~al\mbox{.}}{2017b}]%
        {chen2017improving}
\bibfield{author}{\bibinfo{person}{Tao Chen}, \bibinfo{person}{Ruifeng Xu},
  \bibinfo{person}{Yulan He}, {and} \bibinfo{person}{Xuan Wang}.}
  \bibinfo{year}{2017}\natexlab{b}.
\newblock \showarticletitle{Improving sentiment analysis via sentence type
  classification using BiLSTM-CRF and CNN}.
\newblock \bibinfo{journal}{\emph{Expert Systems with Applications}}
  \bibinfo{volume}{72} (\bibinfo{year}{2017}), \bibinfo{pages}{221--230}.
\newblock


\bibitem[\protect\citeauthoryear{Chen, Wang, and Liu}{Chen
  et~al\mbox{.}}{2017a}]%
        {chen2017visual}
\bibfield{author}{\bibinfo{person}{Xingyue Chen}, \bibinfo{person}{Yunhong
  Wang}, {and} \bibinfo{person}{Qingjie Liu}.}
  \bibinfo{year}{2017}\natexlab{a}.
\newblock \showarticletitle{Visual and textual sentiment analysis using deep
  fusion convolutional neural networks}. In \bibinfo{booktitle}{\emph{2017 IEEE
  International Conference on Image Processing (ICIP)}}. IEEE,
  \bibinfo{pages}{1557--1561}.
\newblock


\bibitem[\protect\citeauthoryear{Correia, Li, and Rocha}{Correia
  et~al\mbox{.}}{2016}]%
        {correia2016monitoring}
\bibfield{author}{\bibinfo{person}{Rion~Brattig Correia}, \bibinfo{person}{Lang
  Li}, {and} \bibinfo{person}{Luis~M Rocha}.} \bibinfo{year}{2016}\natexlab{}.
\newblock \showarticletitle{Monitoring potential drug interactions and
  reactions via network analysis of instagram user timelines}. In
  \bibinfo{booktitle}{\emph{Biocomputing 2016: Proceedings of the Pacific
  Symposium}}. World Scientific, \bibinfo{pages}{492--503}.
\newblock


\bibitem[\protect\citeauthoryear{Cover}{Cover}{1999}]%
        {cover1999elements}
\bibfield{author}{\bibinfo{person}{Thomas~M Cover}.}
  \bibinfo{year}{1999}\natexlab{}.
\newblock \bibinfo{booktitle}{\emph{Elements of information theory}}.
\newblock \bibinfo{publisher}{John Wiley \& Sons}.
\newblock


\bibitem[\protect\citeauthoryear{Devlin, Chang, Lee, and Toutanova}{Devlin
  et~al\mbox{.}}{2018}]%
        {devlin2018bert}
\bibfield{author}{\bibinfo{person}{Jacob Devlin}, \bibinfo{person}{Ming-Wei
  Chang}, \bibinfo{person}{Kenton Lee}, {and} \bibinfo{person}{Kristina
  Toutanova}.} \bibinfo{year}{2018}\natexlab{}.
\newblock \showarticletitle{Bert: Pre-training of deep bidirectional
  transformers for language understanding}.
\newblock \bibinfo{journal}{\emph{arXiv preprint arXiv:1810.04805}}
  (\bibinfo{year}{2018}).
\newblock


\bibitem[\protect\citeauthoryear{Fan, Zhang, Ye, and Li}{Fan
  et~al\mbox{.}}{2018}]%
        {fan2018automatic}
\bibfield{author}{\bibinfo{person}{Yujie Fan}, \bibinfo{person}{Yiming Zhang},
  \bibinfo{person}{Yanfang Ye}, {and} \bibinfo{person}{Xin Li}.}
  \bibinfo{year}{2018}\natexlab{}.
\newblock \showarticletitle{Automatic Opioid User Detection from Twitter:
  Transductive Ensemble Built on Different Meta-graph Based Similarities over
  Heterogeneous Information Network.} \bibinfo{pages}{3357--3363}.
\newblock


\bibitem[\protect\citeauthoryear{Fan, Zhang, Ye, Li, and Zheng}{Fan
  et~al\mbox{.}}{2017}]%
        {fan2017social}
\bibfield{author}{\bibinfo{person}{Yujie Fan}, \bibinfo{person}{Yiming Zhang},
  \bibinfo{person}{Yanfang Ye}, \bibinfo{person}{Xin Li}, {and}
  \bibinfo{person}{Wanhong Zheng}.} \bibinfo{year}{2017}\natexlab{}.
\newblock \showarticletitle{Social media for opioid addiction epidemiology:
  Automatic detection of opioid addicts from twitter and case studies}. In
  \bibinfo{booktitle}{\emph{Proceedings of the 2017 ACM on Conference on
  Information and Knowledge Management}}. \bibinfo{pages}{1259--1267}.
\newblock


\bibitem[\protect\citeauthoryear{Farnadi, Tang, De~Cock, and Moens}{Farnadi
  et~al\mbox{.}}{2018}]%
        {farnadi2018user}
\bibfield{author}{\bibinfo{person}{Golnoosh Farnadi}, \bibinfo{person}{Jie
  Tang}, \bibinfo{person}{Martine De~Cock}, {and}
  \bibinfo{person}{Marie-Francine Moens}.} \bibinfo{year}{2018}\natexlab{}.
\newblock \showarticletitle{User profiling through deep multimodal fusion}. In
  \bibinfo{booktitle}{\emph{Proceedings of the Eleventh ACM International
  Conference on Web Search and Data Mining}}. \bibinfo{pages}{171--179}.
\newblock


\bibitem[\protect\citeauthoryear{Fortunato}{Fortunato}{2010}]%
        {fortunato2010community}
\bibfield{author}{\bibinfo{person}{Santo Fortunato}.}
  \bibinfo{year}{2010}\natexlab{}.
\newblock \showarticletitle{Community detection in graphs}.
\newblock \bibinfo{journal}{\emph{Physics reports}} \bibinfo{volume}{486},
  \bibinfo{number}{3-5} (\bibinfo{year}{2010}), \bibinfo{pages}{75--174}.
\newblock


\bibitem[\protect\citeauthoryear{Fukui, Park, Yang, Rohrbach, Darrell, and
  Rohrbach}{Fukui et~al\mbox{.}}{2016}]%
        {fukui2016multimodal}
\bibfield{author}{\bibinfo{person}{Akira Fukui}, \bibinfo{person}{Dong~Huk
  Park}, \bibinfo{person}{Daylen Yang}, \bibinfo{person}{Anna Rohrbach},
  \bibinfo{person}{Trevor Darrell}, {and} \bibinfo{person}{Marcus Rohrbach}.}
  \bibinfo{year}{2016}\natexlab{}.
\newblock \showarticletitle{Multimodal compact bilinear pooling for visual
  question answering and visual grounding}.
\newblock \bibinfo{journal}{\emph{arXiv preprint arXiv:1606.01847}}
  (\bibinfo{year}{2016}).
\newblock


\bibitem[\protect\citeauthoryear{Gao, Li, Chen, and Zhang}{Gao
  et~al\mbox{.}}{2020a}]%
        {gao2020survey}
\bibfield{author}{\bibinfo{person}{Jing Gao}, \bibinfo{person}{Peng Li},
  \bibinfo{person}{Zhikui Chen}, {and} \bibinfo{person}{Jianing Zhang}.}
  \bibinfo{year}{2020}\natexlab{a}.
\newblock \showarticletitle{A survey on deep learning for multimodal data
  fusion}.
\newblock \bibinfo{journal}{\emph{Neural Computation}} \bibinfo{volume}{32},
  \bibinfo{number}{5} (\bibinfo{year}{2020}), \bibinfo{pages}{829--864}.
\newblock


\bibitem[\protect\citeauthoryear{Gao, Sang, Ren, and Xu}{Gao
  et~al\mbox{.}}{2017}]%
        {gao2017hashtag}
\bibfield{author}{\bibinfo{person}{Yuqi Gao}, \bibinfo{person}{Jitao Sang},
  \bibinfo{person}{Tongwei Ren}, {and} \bibinfo{person}{Changsheng Xu}.}
  \bibinfo{year}{2017}\natexlab{}.
\newblock \showarticletitle{Hashtag-centric immersive search on social media}.
  In \bibinfo{booktitle}{\emph{Proceedings of the 25th ACM international
  conference on Multimedia}}. \bibinfo{pages}{1924--1932}.
\newblock


\bibitem[\protect\citeauthoryear{Gao, Wu, Zhang, Dai, Jia, and Harandi}{Gao
  et~al\mbox{.}}{2020b}]%
        {gao2020revisiting}
\bibfield{author}{\bibinfo{person}{Zhi Gao}, \bibinfo{person}{Yuwei Wu},
  \bibinfo{person}{Xiaoxun Zhang}, \bibinfo{person}{Jindou Dai},
  \bibinfo{person}{Yunde Jia}, {and} \bibinfo{person}{Mehrtash Harandi}.}
  \bibinfo{year}{2020}\natexlab{b}.
\newblock \showarticletitle{Revisiting Bilinear Pooling: A Coding
  Perspective.}. In \bibinfo{booktitle}{\emph{AAAI}}.
  \bibinfo{pages}{3954--3961}.
\newblock


\bibitem[\protect\citeauthoryear{Godin, Slavkovikj, De~Neve, Schrauwen, and
  Van~de Walle}{Godin et~al\mbox{.}}{2013}]%
        {godin2013using}
\bibfield{author}{\bibinfo{person}{Fr{\'e}deric Godin}, \bibinfo{person}{Viktor
  Slavkovikj}, \bibinfo{person}{Wesley De~Neve}, \bibinfo{person}{Benjamin
  Schrauwen}, {and} \bibinfo{person}{Rik Van~de Walle}.}
  \bibinfo{year}{2013}\natexlab{}.
\newblock \showarticletitle{Using topic models for twitter hashtag
  recommendation}. In \bibinfo{booktitle}{\emph{Proceedings of the 22nd
  International Conference on World Wide Web}}. \bibinfo{pages}{593--596}.
\newblock


\bibitem[\protect\citeauthoryear{Goodfellow, Bengio, and Courville}{Goodfellow
  et~al\mbox{.}}{2016}]%
        {goodfellow2016deep}
\bibfield{author}{\bibinfo{person}{Ian Goodfellow}, \bibinfo{person}{Yoshua
  Bengio}, {and} \bibinfo{person}{Aaron Courville}.}
  \bibinfo{year}{2016}\natexlab{}.
\newblock \bibinfo{booktitle}{\emph{Deep learning}}.
\newblock \bibinfo{publisher}{MIT press}.
\newblock


\bibitem[\protect\citeauthoryear{Greenwood, Perrin, and Duggan}{Greenwood
  et~al\mbox{.}}{2016}]%
        {greenwood2016social}
\bibfield{author}{\bibinfo{person}{Shannon Greenwood}, \bibinfo{person}{Andrew
  Perrin}, {and} \bibinfo{person}{Maeve Duggan}.}
  \bibinfo{year}{2016}\natexlab{}.
\newblock \showarticletitle{Social media update 2016}.
\newblock \bibinfo{journal}{\emph{Pew Research Center}} \bibinfo{volume}{11},
  \bibinfo{number}{2} (\bibinfo{year}{2016}).
\newblock


\bibitem[\protect\citeauthoryear{Gunatilaka and Baertlein}{Gunatilaka and
  Baertlein}{2001}]%
        {gunatilaka2001feature}
\bibfield{author}{\bibinfo{person}{Ajith~H. Gunatilaka} {and}
  \bibinfo{person}{Brian~A. Baertlein}.} \bibinfo{year}{2001}\natexlab{}.
\newblock \showarticletitle{Feature-level and decision-level fusion of
  noncoincidently sampled sensors for land mine detection}.
\newblock \bibinfo{journal}{\emph{IEEE transactions on pattern analysis and
  machine intelligence}} \bibinfo{volume}{23}, \bibinfo{number}{6}
  (\bibinfo{year}{2001}), \bibinfo{pages}{577--589}.
\newblock


\bibitem[\protect\citeauthoryear{Guy, Zwerdling, Ronen, Carmel, and Uziel}{Guy
  et~al\mbox{.}}{2010}]%
        {guy2010social}
\bibfield{author}{\bibinfo{person}{Ido Guy}, \bibinfo{person}{Naama Zwerdling},
  \bibinfo{person}{Inbal Ronen}, \bibinfo{person}{David Carmel}, {and}
  \bibinfo{person}{Erel Uziel}.} \bibinfo{year}{2010}\natexlab{}.
\newblock \showarticletitle{Social media recommendation based on people and
  tags}. In \bibinfo{booktitle}{\emph{Proceedings of the 33rd international ACM
  SIGIR conference on Research and development in information retrieval}}.
  \bibinfo{pages}{194--201}.
\newblock


\bibitem[\protect\citeauthoryear{Hagberg and Conway}{Hagberg and
  Conway}{2010}]%
        {hagberg2010hacking}
\bibfield{author}{\bibinfo{person}{A Hagberg} {and} \bibinfo{person}{D
  Conway}.} \bibinfo{year}{2010}\natexlab{}.
\newblock \showarticletitle{Hacking social networks using the Python
  programming language}.
\newblock \bibinfo{journal}{\emph{Sunbelt 2010, Riva del Garda, Italy}}
  (\bibinfo{year}{2010}).
\newblock


\bibitem[\protect\citeauthoryear{Hagberg, Swart, and S~Chult}{Hagberg
  et~al\mbox{.}}{2008}]%
        {hagberg2008exploring}
\bibfield{author}{\bibinfo{person}{Aric Hagberg}, \bibinfo{person}{Pieter
  Swart}, {and} \bibinfo{person}{Daniel S~Chult}.}
  \bibinfo{year}{2008}\natexlab{}.
\newblock \bibinfo{booktitle}{\emph{Exploring network structure, dynamics, and
  function using NetworkX}}.
\newblock \bibinfo{type}{{T}echnical {R}eport}. \bibinfo{institution}{Los
  Alamos National Lab.(LANL), Los Alamos, NM (United States)}.
\newblock


\bibitem[\protect\citeauthoryear{Haghighat, Abdel-Mottaleb, and
  Alhalabi}{Haghighat et~al\mbox{.}}{2016}]%
        {haghighat2016discriminant}
\bibfield{author}{\bibinfo{person}{Mohammad Haghighat},
  \bibinfo{person}{Mohamed Abdel-Mottaleb}, {and} \bibinfo{person}{Wadee
  Alhalabi}.} \bibinfo{year}{2016}\natexlab{}.
\newblock \showarticletitle{Discriminant correlation analysis: Real-time
  feature level fusion for multimodal biometric recognition}.
\newblock \bibinfo{journal}{\emph{IEEE Transactions on Information Forensics
  and Security}} \bibinfo{volume}{11}, \bibinfo{number}{9}
  (\bibinfo{year}{2016}), \bibinfo{pages}{1984--1996}.
\newblock


\bibitem[\protect\citeauthoryear{Hassanpour, Tomita, DeLise, Crosier, and
  Marsch}{Hassanpour et~al\mbox{.}}{2019}]%
        {hassanpour2019identifying}
\bibfield{author}{\bibinfo{person}{Saeed Hassanpour}, \bibinfo{person}{Naofumi
  Tomita}, \bibinfo{person}{Timothy DeLise}, \bibinfo{person}{Benjamin
  Crosier}, {and} \bibinfo{person}{Lisa~A Marsch}.}
  \bibinfo{year}{2019}\natexlab{}.
\newblock \showarticletitle{Identifying substance use risk based on deep neural
  networks and Instagram social media data}.
\newblock \bibinfo{journal}{\emph{Neuropsychopharmacology}}
  \bibinfo{volume}{44}, \bibinfo{number}{3} (\bibinfo{year}{2019}),
  \bibinfo{pages}{487--494}.
\newblock


\bibitem[\protect\citeauthoryear{He, Zhang, Ren, and Sun}{He
  et~al\mbox{.}}{2016}]%
        {he2016deep}
\bibfield{author}{\bibinfo{person}{Kaiming He}, \bibinfo{person}{Xiangyu
  Zhang}, \bibinfo{person}{Shaoqing Ren}, {and} \bibinfo{person}{Jian Sun}.}
  \bibinfo{year}{2016}\natexlab{}.
\newblock \showarticletitle{Deep residual learning for image recognition}. In
  \bibinfo{booktitle}{\emph{Proceedings of the IEEE conference on computer
  vision and pattern recognition}}. \bibinfo{pages}{770--778}.
\newblock


\bibitem[\protect\citeauthoryear{Hu, Phan, Chun, Geller, Vo, Ye, Jin, Ding,
  Kenne, and Dou}{Hu et~al\mbox{.}}{2019}]%
        {hu2019insight}
\bibfield{author}{\bibinfo{person}{Han Hu}, \bibinfo{person}{NhatHai Phan},
  \bibinfo{person}{Soon~A Chun}, \bibinfo{person}{James Geller},
  \bibinfo{person}{Huy Vo}, \bibinfo{person}{Xinyue Ye},
  \bibinfo{person}{Ruoming Jin}, \bibinfo{person}{Kele Ding},
  \bibinfo{person}{Deric Kenne}, {and} \bibinfo{person}{Dejing Dou}.}
  \bibinfo{year}{2019}\natexlab{}.
\newblock \showarticletitle{An insight analysis and detection of drug-abuse
  risk behavior on Twitter with self-taught deep learning}.
\newblock \bibinfo{journal}{\emph{Computational Social Networks}}
  \bibinfo{volume}{6}, \bibinfo{number}{1} (\bibinfo{year}{2019}),
  \bibinfo{pages}{10}.
\newblock


\bibitem[\protect\citeauthoryear{Hu, Manikonda, and Kambhampati}{Hu
  et~al\mbox{.}}{2014}]%
        {hu2014we}
\bibfield{author}{\bibinfo{person}{Yuheng Hu}, \bibinfo{person}{Lydia
  Manikonda}, {and} \bibinfo{person}{Subbarao Kambhampati}.}
  \bibinfo{year}{2014}\natexlab{}.
\newblock \showarticletitle{What we instagram: A first analysis of instagram
  photo content and user types}. In \bibinfo{booktitle}{\emph{Eighth
  International AAAI conference on weblogs and social media}}.
\newblock


\bibitem[\protect\citeauthoryear{Huang, Liu, Van Der~Maaten, and
  Weinberger}{Huang et~al\mbox{.}}{2017}]%
        {huang2017densely}
\bibfield{author}{\bibinfo{person}{Gao Huang}, \bibinfo{person}{Zhuang Liu},
  \bibinfo{person}{Laurens Van Der~Maaten}, {and} \bibinfo{person}{Kilian~Q
  Weinberger}.} \bibinfo{year}{2017}\natexlab{}.
\newblock \showarticletitle{Densely connected convolutional networks}. In
  \bibinfo{booktitle}{\emph{Proceedings of the IEEE conference on computer
  vision and pattern recognition}}. \bibinfo{pages}{4700--4708}.
\newblock


\bibitem[\protect\citeauthoryear{Hughes, Bright, and Chalmers}{Hughes
  et~al\mbox{.}}{2017}]%
        {hughes2017social}
\bibfield{author}{\bibinfo{person}{Caitlin~E Hughes}, \bibinfo{person}{David~A
  Bright}, {and} \bibinfo{person}{Jenny Chalmers}.}
  \bibinfo{year}{2017}\natexlab{}.
\newblock \showarticletitle{Social network analysis of Australian poly-drug
  trafficking networks: How do drug traffickers manage multiple illicit drugs?}
\newblock \bibinfo{journal}{\emph{Social Networks}}  \bibinfo{volume}{51}
  (\bibinfo{year}{2017}), \bibinfo{pages}{135--147}.
\newblock


\bibitem[\protect\citeauthoryear{Hung, Huang, Hsu, and Wu}{Hung
  et~al\mbox{.}}{2008}]%
        {hung2008tag}
\bibfield{author}{\bibinfo{person}{Chia-Chuan Hung}, \bibinfo{person}{Yi-Ching
  Huang}, \bibinfo{person}{Jane Yung-jen Hsu}, {and} \bibinfo{person}{David
  Kuan-Chun Wu}.} \bibinfo{year}{2008}\natexlab{}.
\newblock \showarticletitle{Tag-based user profiling for social media
  recommendation}. In \bibinfo{booktitle}{\emph{Workshop on Intelligent
  Techniques for Web Personalization \& Recommender Systems at AAAI}},
  Vol.~\bibinfo{volume}{8}. \bibinfo{pages}{49--55}.
\newblock


\bibitem[\protect\citeauthoryear{Ikeda, Hattori, Ono, Asoh, and
  Higashino}{Ikeda et~al\mbox{.}}{2013}]%
        {ikeda2013twitter}
\bibfield{author}{\bibinfo{person}{Kazushi Ikeda}, \bibinfo{person}{Gen
  Hattori}, \bibinfo{person}{Chihiro Ono}, \bibinfo{person}{Hideki Asoh}, {and}
  \bibinfo{person}{Teruo Higashino}.} \bibinfo{year}{2013}\natexlab{}.
\newblock \showarticletitle{Twitter user profiling based on text and community
  mining for market analysis}.
\newblock \bibinfo{journal}{\emph{Knowledge-Based Systems}}
  \bibinfo{volume}{51} (\bibinfo{year}{2013}), \bibinfo{pages}{35--47}.
\newblock


\bibitem[\protect\citeauthoryear{Jang, Han, Shih, and Lee}{Jang
  et~al\mbox{.}}{2015}]%
        {jang2015generation}
\bibfield{author}{\bibinfo{person}{Jin~Yea Jang}, \bibinfo{person}{Kyungsik
  Han}, \bibinfo{person}{Patrick~C Shih}, {and} \bibinfo{person}{Dongwon Lee}.}
  \bibinfo{year}{2015}\natexlab{}.
\newblock \showarticletitle{Generation like: Comparative characteristics in
  instagram}. In \bibinfo{booktitle}{\emph{Proceedings of the 33rd Annual ACM
  Conference on Human Factors in Computing Systems}}.
  \bibinfo{pages}{4039--4042}.
\newblock


\bibitem[\protect\citeauthoryear{Jia, Hu, Guo, and Xu}{Jia
  et~al\mbox{.}}{2019}]%
        {jia2019database}
\bibfield{author}{\bibinfo{person}{Shan Jia}, \bibinfo{person}{Chuanbo Hu},
  \bibinfo{person}{Guodong Guo}, {and} \bibinfo{person}{Zhengquan Xu}.}
  \bibinfo{year}{2019}\natexlab{}.
\newblock \showarticletitle{A database for face presentation attack using wax
  figure faces}. In \bibinfo{booktitle}{\emph{International Conference on Image
  Analysis and Processing}}. Springer, \bibinfo{pages}{39--47}.
\newblock


\bibitem[\protect\citeauthoryear{Jia, Li, Hu, Guo, and Xu}{Jia
  et~al\mbox{.}}{2020}]%
        {jia20203d}
\bibfield{author}{\bibinfo{person}{Shan Jia}, \bibinfo{person}{Xin Li},
  \bibinfo{person}{Chuanbo Hu}, \bibinfo{person}{Guodong Guo}, {and}
  \bibinfo{person}{Zhengquan Xu}.} \bibinfo{year}{2020}\natexlab{}.
\newblock \showarticletitle{3D Face Anti-Spoofing with Factorized Bilinear
  Coding}.
\newblock \bibinfo{journal}{\emph{IEEE Transactions on Circuits and Systems for
  Video Technology}} (\bibinfo{year}{2020}).
\newblock


\bibitem[\protect\citeauthoryear{Kalchbrenner, Grefenstette, and
  Blunsom}{Kalchbrenner et~al\mbox{.}}{2014}]%
        {kalchbrenner2014convolutional}
\bibfield{author}{\bibinfo{person}{Nal Kalchbrenner}, \bibinfo{person}{Edward
  Grefenstette}, {and} \bibinfo{person}{Phil Blunsom}.}
  \bibinfo{year}{2014}\natexlab{}.
\newblock \showarticletitle{A convolutional neural network for modelling
  sentences}.
\newblock \bibinfo{journal}{\emph{arXiv preprint arXiv:1404.2188}}
  (\bibinfo{year}{2014}).
\newblock


\bibitem[\protect\citeauthoryear{Kalyanam and Mackey}{Kalyanam and
  Mackey}{2017}]%
        {kalyanam2017review}
\bibfield{author}{\bibinfo{person}{Janani Kalyanam} {and}
  \bibinfo{person}{Tim~K Mackey}.} \bibinfo{year}{2017}\natexlab{}.
\newblock \showarticletitle{A review of digital surveillance methods and
  approaches to combat prescription drug abuse}.
\newblock \bibinfo{journal}{\emph{Current Addiction Reports}}
  \bibinfo{volume}{4}, \bibinfo{number}{4} (\bibinfo{year}{2017}),
  \bibinfo{pages}{397--409}.
\newblock


\bibitem[\protect\citeauthoryear{Kilmer, Sohler~Everingham, Caulkins, Midgette,
  Pacula, Reuter, Burns, Han, and Lundberg}{Kilmer et~al\mbox{.}}{2014}]%
        {kilmer2014big}
\bibfield{author}{\bibinfo{person}{Beau Kilmer}, \bibinfo{person}{Susan~S
  Sohler~Everingham}, \bibinfo{person}{Jonathan~P Caulkins},
  \bibinfo{person}{Gregory Midgette}, \bibinfo{person}{Rosalie~Liccardo
  Pacula}, \bibinfo{person}{Peter Reuter}, \bibinfo{person}{Rachel~M Burns},
  \bibinfo{person}{Bing Han}, {and} \bibinfo{person}{Russell Lundberg}.}
  \bibinfo{year}{2014}\natexlab{}.
\newblock \showarticletitle{How big is the US market for illegal drugs?}
\newblock  (\bibinfo{year}{2014}).
\newblock


\bibitem[\protect\citeauthoryear{Kim}{Kim}{2014}]%
        {kim2014convolutional}
\bibfield{author}{\bibinfo{person}{Yoon Kim}.} \bibinfo{year}{2014}\natexlab{}.
\newblock \showarticletitle{Convolutional neural networks for sentence
  classification}.
\newblock \bibinfo{journal}{\emph{arXiv preprint arXiv:1408.5882}}
  (\bibinfo{year}{2014}).
\newblock


\bibitem[\protect\citeauthoryear{Lahat, Adali, and Jutten}{Lahat
  et~al\mbox{.}}{2015}]%
        {lahat2015multimodal}
\bibfield{author}{\bibinfo{person}{Dana Lahat}, \bibinfo{person}{T{\"u}lay
  Adali}, {and} \bibinfo{person}{Christian Jutten}.}
  \bibinfo{year}{2015}\natexlab{}.
\newblock \showarticletitle{Multimodal data fusion: an overview of methods,
  challenges, and prospects}.
\newblock \bibinfo{journal}{\emph{Proc. IEEE}} \bibinfo{volume}{103},
  \bibinfo{number}{9} (\bibinfo{year}{2015}), \bibinfo{pages}{1449--1477}.
\newblock


\bibitem[\protect\citeauthoryear{Lancichinetti and Fortunato}{Lancichinetti and
  Fortunato}{2009}]%
        {lancichinetti2009community}
\bibfield{author}{\bibinfo{person}{Andrea Lancichinetti} {and}
  \bibinfo{person}{Santo Fortunato}.} \bibinfo{year}{2009}\natexlab{}.
\newblock \showarticletitle{Community detection algorithms: a comparative
  analysis}.
\newblock \bibinfo{journal}{\emph{Physical review E}} \bibinfo{volume}{80},
  \bibinfo{number}{5} (\bibinfo{year}{2009}), \bibinfo{pages}{056117}.
\newblock


\bibitem[\protect\citeauthoryear{Li, Xu, Shah, and Mackey}{Li
  et~al\mbox{.}}{2019}]%
        {li2019machine}
\bibfield{author}{\bibinfo{person}{Jiawei Li}, \bibinfo{person}{Qing Xu},
  \bibinfo{person}{Neal Shah}, {and} \bibinfo{person}{Tim~K Mackey}.}
  \bibinfo{year}{2019}\natexlab{}.
\newblock \showarticletitle{A machine learning approach for the detection and
  characterization of illicit drug dealers on instagram: model evaluation
  study}.
\newblock \bibinfo{journal}{\emph{Journal of medical Internet research}}
  \bibinfo{volume}{21}, \bibinfo{number}{6} (\bibinfo{year}{2019}),
  \bibinfo{pages}{e13803}.
\newblock


\bibitem[\protect\citeauthoryear{Li, Wang, Deng, Wang, and Chang}{Li
  et~al\mbox{.}}{2012}]%
        {li2012towards}
\bibfield{author}{\bibinfo{person}{Rui Li}, \bibinfo{person}{Shengjie Wang},
  \bibinfo{person}{Hongbo Deng}, \bibinfo{person}{Rui Wang}, {and}
  \bibinfo{person}{Kevin Chen-Chuan Chang}.} \bibinfo{year}{2012}\natexlab{}.
\newblock \showarticletitle{Towards social user profiling: unified and
  discriminative influence model for inferring home locations}. In
  \bibinfo{booktitle}{\emph{Proceedings of the 18th ACM SIGKDD international
  conference on Knowledge discovery and data mining}}.
  \bibinfo{pages}{1023--1031}.
\newblock


\bibitem[\protect\citeauthoryear{Li, Wu, Ester, Kao, Wang, and Zheng}{Li
  et~al\mbox{.}}{2017}]%
        {li2017semi}
\bibfield{author}{\bibinfo{person}{Xiang Li}, \bibinfo{person}{Yao Wu},
  \bibinfo{person}{Martin Ester}, \bibinfo{person}{Ben Kao},
  \bibinfo{person}{Xin Wang}, {and} \bibinfo{person}{Yudian Zheng}.}
  \bibinfo{year}{2017}\natexlab{}.
\newblock \showarticletitle{Semi-supervised clustering in attributed
  heterogeneous information networks}. In \bibinfo{booktitle}{\emph{Proceedings
  of the 26th International Conference on World Wide Web}}.
  \bibinfo{pages}{1621--1629}.
\newblock


\bibitem[\protect\citeauthoryear{Lin, RoyChowdhury, and Maji}{Lin
  et~al\mbox{.}}{2015}]%
        {lin2015bilinear}
\bibfield{author}{\bibinfo{person}{Tsung-Yu Lin}, \bibinfo{person}{Aruni
  RoyChowdhury}, {and} \bibinfo{person}{Subhransu Maji}.}
  \bibinfo{year}{2015}\natexlab{}.
\newblock \showarticletitle{Bilinear cnn models for fine-grained visual
  recognition}. In \bibinfo{booktitle}{\emph{Proceedings of the IEEE
  international conference on computer vision}}. \bibinfo{pages}{1449--1457}.
\newblock


\bibitem[\protect\citeauthoryear{Ling, Lyu, and King}{Ling
  et~al\mbox{.}}{2014}]%
        {ling2014ratings}
\bibfield{author}{\bibinfo{person}{Guang Ling}, \bibinfo{person}{Michael~R
  Lyu}, {and} \bibinfo{person}{Irwin King}.} \bibinfo{year}{2014}\natexlab{}.
\newblock \showarticletitle{Ratings meet reviews, a combined approach to
  recommend}. In \bibinfo{booktitle}{\emph{Proceedings of the 8th ACM
  Conference on Recommender systems}}. \bibinfo{pages}{105--112}.
\newblock


\bibitem[\protect\citeauthoryear{Liu, Zhang, Lin, and Quek}{Liu
  et~al\mbox{.}}{2017}]%
        {liu2017heterogeneous}
\bibfield{author}{\bibinfo{person}{Zuozhu Liu}, \bibinfo{person}{Wenyu Zhang},
  \bibinfo{person}{Shaowei Lin}, {and} \bibinfo{person}{Tony~QS Quek}.}
  \bibinfo{year}{2017}\natexlab{}.
\newblock \showarticletitle{Heterogeneous sensor data fusion by deep multimodal
  encoding}.
\newblock \bibinfo{journal}{\emph{IEEE Journal of Selected Topics in Signal
  Processing}} \bibinfo{volume}{11}, \bibinfo{number}{3}
  (\bibinfo{year}{2017}), \bibinfo{pages}{479--491}.
\newblock


\bibitem[\protect\citeauthoryear{Luxton, June, and Fairall}{Luxton
  et~al\mbox{.}}{2012}]%
        {luxton2012social}
\bibfield{author}{\bibinfo{person}{David~D Luxton}, \bibinfo{person}{Jennifer~D
  June}, {and} \bibinfo{person}{Jonathan~M Fairall}.}
  \bibinfo{year}{2012}\natexlab{}.
\newblock \showarticletitle{Social media and suicide: a public health
  perspective}.
\newblock \bibinfo{journal}{\emph{American journal of public health}}
  \bibinfo{volume}{102}, \bibinfo{number}{S2} (\bibinfo{year}{2012}),
  \bibinfo{pages}{S195--S200}.
\newblock


\bibitem[\protect\citeauthoryear{Ma, Ren, Zhao, Tulyakov, Wu, and Peng}{Ma
  et~al\mbox{.}}{2021}]%
        {ma2021smil}
\bibfield{author}{\bibinfo{person}{Mengmeng Ma}, \bibinfo{person}{Jian Ren},
  \bibinfo{person}{Long Zhao}, \bibinfo{person}{Sergey Tulyakov},
  \bibinfo{person}{Cathy Wu}, {and} \bibinfo{person}{Xi Peng}.}
  \bibinfo{year}{2021}\natexlab{}.
\newblock \showarticletitle{SMIL: Multimodal Learning with Severely Missing
  Modality}.
\newblock \bibinfo{journal}{\emph{arXiv preprint arXiv:2103.05677}}
  (\bibinfo{year}{2021}).
\newblock


\bibitem[\protect\citeauthoryear{Mackey, Kalyanam, Klugman, Kuzmenko, and
  Gupta}{Mackey et~al\mbox{.}}{2018}]%
        {mackey2018solution}
\bibfield{author}{\bibinfo{person}{Tim Mackey}, \bibinfo{person}{Janani
  Kalyanam}, \bibinfo{person}{Josh Klugman}, \bibinfo{person}{Ella Kuzmenko},
  {and} \bibinfo{person}{Rashmi Gupta}.} \bibinfo{year}{2018}\natexlab{}.
\newblock \showarticletitle{Solution to detect, classify, and report illicit
  online marketing and sales of controlled substances via Twitter: using
  machine learning and web forensics to combat digital opioid access}.
\newblock \bibinfo{journal}{\emph{Journal of medical Internet research}}
  \bibinfo{volume}{20}, \bibinfo{number}{4} (\bibinfo{year}{2018}),
  \bibinfo{pages}{e10029}.
\newblock


\bibitem[\protect\citeauthoryear{Maeda}{Maeda}{2012}]%
        {maeda2012performance}
\bibfield{author}{\bibinfo{person}{Kazuaki Maeda}.}
  \bibinfo{year}{2012}\natexlab{}.
\newblock \showarticletitle{Performance evaluation of object serialization
  libraries in XML, JSON and binary formats}. In \bibinfo{booktitle}{\emph{2012
  Second International Conference on Digital Information and Communication
  Technology and it's Applications (DICTAP)}}. IEEE, \bibinfo{pages}{177--182}.
\newblock


\bibitem[\protect\citeauthoryear{Newman}{Newman}{2006}]%
        {newman2006modularity}
\bibfield{author}{\bibinfo{person}{Mark~EJ Newman}.}
  \bibinfo{year}{2006}\natexlab{}.
\newblock \showarticletitle{Modularity and community structure in networks}.
\newblock \bibinfo{journal}{\emph{Proceedings of the national academy of
  sciences}} \bibinfo{volume}{103}, \bibinfo{number}{23}
  (\bibinfo{year}{2006}), \bibinfo{pages}{8577--8582}.
\newblock


\bibitem[\protect\citeauthoryear{Newman and Girvan}{Newman and Girvan}{2004}]%
        {newman2004finding}
\bibfield{author}{\bibinfo{person}{Mark~EJ Newman} {and}
  \bibinfo{person}{Michelle Girvan}.} \bibinfo{year}{2004}\natexlab{}.
\newblock \showarticletitle{Finding and evaluating community structure in
  networks}.
\newblock \bibinfo{journal}{\emph{Physical review E}} \bibinfo{volume}{69},
  \bibinfo{number}{2} (\bibinfo{year}{2004}), \bibinfo{pages}{026113}.
\newblock


\bibitem[\protect\citeauthoryear{Nie, Song, and Chua}{Nie
  et~al\mbox{.}}{2016}]%
        {nie2016learning}
\bibfield{author}{\bibinfo{person}{Liqiang Nie}, \bibinfo{person}{Xuemeng
  Song}, {and} \bibinfo{person}{Tat-Seng Chua}.}
  \bibinfo{year}{2016}\natexlab{}.
\newblock \showarticletitle{Learning from multiple social networks}.
\newblock \bibinfo{journal}{\emph{Synthesis lectures on information concepts,
  retrieval, and services}} \bibinfo{volume}{8}, \bibinfo{number}{2}
  (\bibinfo{year}{2016}), \bibinfo{pages}{1--118}.
\newblock


\bibitem[\protect\citeauthoryear{O'Connor, Balasubramanyan, Routledge, and
  Smith}{O'Connor et~al\mbox{.}}{2010}]%
        {oConnor2010tweets}
\bibfield{author}{\bibinfo{person}{Brendan O'Connor}, \bibinfo{person}{Ramnath
  Balasubramanyan}, \bibinfo{person}{Bryan~R Routledge}, {and}
  \bibinfo{person}{Noah~A Smith}.} \bibinfo{year}{2010}\natexlab{}.
\newblock \showarticletitle{From tweets to polls: Linking text sentiment to
  public opinion time series}. In \bibinfo{booktitle}{\emph{Fourth
  international AAAI conference on weblogs and social media}}.
\newblock


\bibitem[\protect\citeauthoryear{Olston and Najork}{Olston and Najork}{2010}]%
        {olston2010web}
\bibfield{author}{\bibinfo{person}{Christopher Olston} {and}
  \bibinfo{person}{Marc Najork}.} \bibinfo{year}{2010}\natexlab{}.
\newblock \bibinfo{booktitle}{\emph{Web crawling}}.
\newblock \bibinfo{publisher}{Now Publishers Inc}.
\newblock


\bibitem[\protect\citeauthoryear{Pan and Yang}{Pan and Yang}{2009}]%
        {pan2009survey}
\bibfield{author}{\bibinfo{person}{Sinno~Jialin Pan} {and}
  \bibinfo{person}{Qiang Yang}.} \bibinfo{year}{2009}\natexlab{}.
\newblock \showarticletitle{A survey on transfer learning}.
\newblock \bibinfo{journal}{\emph{IEEE Transactions on knowledge and data
  engineering}} \bibinfo{volume}{22}, \bibinfo{number}{10}
  (\bibinfo{year}{2009}), \bibinfo{pages}{1345--1359}.
\newblock


\bibitem[\protect\citeauthoryear{Papadopoulos, Kompatsiaris, Vakali, and
  Spyridonos}{Papadopoulos et~al\mbox{.}}{2012}]%
        {papadopoulos2012community}
\bibfield{author}{\bibinfo{person}{Symeon Papadopoulos},
  \bibinfo{person}{Yiannis Kompatsiaris}, \bibinfo{person}{Athena Vakali},
  {and} \bibinfo{person}{Ploutarchos Spyridonos}.}
  \bibinfo{year}{2012}\natexlab{}.
\newblock \showarticletitle{Community detection in social media}.
\newblock \bibinfo{journal}{\emph{Data Mining and Knowledge Discovery}}
  \bibinfo{volume}{24}, \bibinfo{number}{3} (\bibinfo{year}{2012}),
  \bibinfo{pages}{515--554}.
\newblock


\bibitem[\protect\citeauthoryear{Poggio, Mhaskar, Rosasco, Miranda, and
  Liao}{Poggio et~al\mbox{.}}{2017}]%
        {poggio2017and}
\bibfield{author}{\bibinfo{person}{Tomaso Poggio}, \bibinfo{person}{Hrushikesh
  Mhaskar}, \bibinfo{person}{Lorenzo Rosasco}, \bibinfo{person}{Brando
  Miranda}, {and} \bibinfo{person}{Qianli Liao}.}
  \bibinfo{year}{2017}\natexlab{}.
\newblock \showarticletitle{Why and when can deep-but not shallow-networks
  avoid the curse of dimensionality: a review}.
\newblock \bibinfo{journal}{\emph{International Journal of Automation and
  Computing}} \bibinfo{volume}{14}, \bibinfo{number}{5} (\bibinfo{year}{2017}),
  \bibinfo{pages}{503--519}.
\newblock


\bibitem[\protect\citeauthoryear{Reddi, Kale, and Kumar}{Reddi
  et~al\mbox{.}}{2019}]%
        {reddi2019convergence}
\bibfield{author}{\bibinfo{person}{Sashank~J Reddi}, \bibinfo{person}{Satyen
  Kale}, {and} \bibinfo{person}{Sanjiv Kumar}.}
  \bibinfo{year}{2019}\natexlab{}.
\newblock \showarticletitle{On the convergence of adam and beyond}.
\newblock \bibinfo{journal}{\emph{arXiv preprint arXiv:1904.09237}}
  (\bibinfo{year}{2019}).
\newblock


\bibitem[\protect\citeauthoryear{Sarker, DeRoos, and Perrone}{Sarker
  et~al\mbox{.}}{2020}]%
        {sarker2020mining}
\bibfield{author}{\bibinfo{person}{Abeed Sarker}, \bibinfo{person}{Annika
  DeRoos}, {and} \bibinfo{person}{Jeanmarie Perrone}.}
  \bibinfo{year}{2020}\natexlab{}.
\newblock \showarticletitle{Mining social media for prescription medication
  abuse monitoring: a review and proposal for a data-centric framework}.
\newblock \bibinfo{journal}{\emph{Journal of the American Medical Informatics
  Association}} \bibinfo{volume}{27}, \bibinfo{number}{2}
  (\bibinfo{year}{2020}), \bibinfo{pages}{315--329}.
\newblock


\bibitem[\protect\citeauthoryear{Sarker, Gonzalez-Hernandez, Ruan, and
  Perrone}{Sarker et~al\mbox{.}}{2019}]%
        {sarker2019machine}
\bibfield{author}{\bibinfo{person}{Abeed Sarker}, \bibinfo{person}{Graciela
  Gonzalez-Hernandez}, \bibinfo{person}{Yucheng Ruan}, {and}
  \bibinfo{person}{Jeanmarie Perrone}.} \bibinfo{year}{2019}\natexlab{}.
\newblock \showarticletitle{Machine learning and natural language processing
  for geolocation-centric monitoring and characterization of opioid-related
  social media chatter}.
\newblock \bibinfo{journal}{\emph{JAMA network open}} \bibinfo{volume}{2},
  \bibinfo{number}{11} (\bibinfo{year}{2019}),
  \bibinfo{pages}{e1914672--e1914672}.
\newblock


\bibitem[\protect\citeauthoryear{Scott}{Scott}{2015}]%
        {scott2015new}
\bibfield{author}{\bibinfo{person}{David~Meerman Scott}.}
  \bibinfo{year}{2015}\natexlab{}.
\newblock \bibinfo{booktitle}{\emph{The new rules of marketing and PR: How to
  use social media, online video, mobile applications, blogs, news releases,
  and viral marketing to reach buyers directly}}.
\newblock \bibinfo{publisher}{John Wiley \& Sons}.
\newblock


\bibitem[\protect\citeauthoryear{Shang, Palmer, Sun, Chen, Lu, and Bi}{Shang
  et~al\mbox{.}}{2017}]%
        {shang2017vigan}
\bibfield{author}{\bibinfo{person}{Chao Shang}, \bibinfo{person}{Aaron Palmer},
  \bibinfo{person}{Jiangwen Sun}, \bibinfo{person}{Ko-Shin Chen},
  \bibinfo{person}{Jin Lu}, {and} \bibinfo{person}{Jinbo Bi}.}
  \bibinfo{year}{2017}\natexlab{}.
\newblock \showarticletitle{VIGAN: Missing view imputation with generative
  adversarial networks}. In \bibinfo{booktitle}{\emph{2017 IEEE International
  Conference on Big Data (Big Data)}}. IEEE, \bibinfo{pages}{766--775}.
\newblock


\bibitem[\protect\citeauthoryear{Shi, Li, Zhang, Sun, and Philip}{Shi
  et~al\mbox{.}}{2016}]%
        {shi2016survey}
\bibfield{author}{\bibinfo{person}{Chuan Shi}, \bibinfo{person}{Yitong Li},
  \bibinfo{person}{Jiawei Zhang}, \bibinfo{person}{Yizhou Sun}, {and}
  \bibinfo{person}{S~Yu Philip}.} \bibinfo{year}{2016}\natexlab{}.
\newblock \showarticletitle{A survey of heterogeneous information network
  analysis}.
\newblock \bibinfo{journal}{\emph{IEEE Transactions on Knowledge and Data
  Engineering}} \bibinfo{volume}{29}, \bibinfo{number}{1}
  (\bibinfo{year}{2016}), \bibinfo{pages}{17--37}.
\newblock


\bibitem[\protect\citeauthoryear{Simonyan and Zisserman}{Simonyan and
  Zisserman}{2014}]%
        {simonyan2014very}
\bibfield{author}{\bibinfo{person}{Karen Simonyan} {and}
  \bibinfo{person}{Andrew Zisserman}.} \bibinfo{year}{2014}\natexlab{}.
\newblock \showarticletitle{Very deep convolutional networks for large-scale
  image recognition}.
\newblock \bibinfo{journal}{\emph{arXiv preprint arXiv:1409.1556}}
  (\bibinfo{year}{2014}).
\newblock


\bibitem[\protect\citeauthoryear{Song, Ming, Nie, Zhao, and Chua}{Song
  et~al\mbox{.}}{2016}]%
        {song2016volunteerism}
\bibfield{author}{\bibinfo{person}{Xuemeng Song}, \bibinfo{person}{Zhao-Yan
  Ming}, \bibinfo{person}{Liqiang Nie}, \bibinfo{person}{Yi-Liang Zhao}, {and}
  \bibinfo{person}{Tat-Seng Chua}.} \bibinfo{year}{2016}\natexlab{}.
\newblock \showarticletitle{Volunteerism tendency prediction via harvesting
  multiple social networks}.
\newblock \bibinfo{journal}{\emph{ACM Transactions on Information Systems
  (TOIS)}} \bibinfo{volume}{34}, \bibinfo{number}{2} (\bibinfo{year}{2016}),
  \bibinfo{pages}{1--27}.
\newblock


\bibitem[\protect\citeauthoryear{Song, Nie, Zhang, Akbari, and Chua}{Song
  et~al\mbox{.}}{2015a}]%
        {song2015multiple}
\bibfield{author}{\bibinfo{person}{Xuemeng Song}, \bibinfo{person}{Liqiang
  Nie}, \bibinfo{person}{Luming Zhang}, \bibinfo{person}{Mohammad Akbari},
  {and} \bibinfo{person}{Tat-Seng Chua}.} \bibinfo{year}{2015}\natexlab{a}.
\newblock \showarticletitle{Multiple social network learning and its
  application in volunteerism tendency prediction}. In
  \bibinfo{booktitle}{\emph{Proceedings of the 38th International ACM SIGIR
  Conference on Research and Development in Information Retrieval}}.
  \bibinfo{pages}{213--222}.
\newblock


\bibitem[\protect\citeauthoryear{Song, Nie, Zhang, Liu, and Chua}{Song
  et~al\mbox{.}}{2015b}]%
        {song2015interest}
\bibfield{author}{\bibinfo{person}{Xuemeng Song}, \bibinfo{person}{Liqiang
  Nie}, \bibinfo{person}{Luming Zhang}, \bibinfo{person}{Maofu Liu}, {and}
  \bibinfo{person}{Tat-Seng Chua}.} \bibinfo{year}{2015}\natexlab{b}.
\newblock \showarticletitle{Interest inference via structure-constrained
  multi-source multi-task learning}. In \bibinfo{booktitle}{\emph{Twenty-Fourth
  International Joint Conference on Artificial Intelligence}}.
\newblock


\bibitem[\protect\citeauthoryear{Stieglitz, Mirbabaie, Ross, and
  Neuberger}{Stieglitz et~al\mbox{.}}{2018}]%
        {stieglitz2018social}
\bibfield{author}{\bibinfo{person}{Stefan Stieglitz}, \bibinfo{person}{Milad
  Mirbabaie}, \bibinfo{person}{Bj{\"o}rn Ross}, {and}
  \bibinfo{person}{Christoph Neuberger}.} \bibinfo{year}{2018}\natexlab{}.
\newblock \showarticletitle{Social media analytics--Challenges in topic
  discovery, data collection, and data preparation}.
\newblock \bibinfo{journal}{\emph{International journal of information
  management}}  \bibinfo{volume}{39} (\bibinfo{year}{2018}),
  \bibinfo{pages}{156--168}.
\newblock


\bibitem[\protect\citeauthoryear{Sun and Han}{Sun and Han}{2012}]%
        {sun2012mining}
\bibfield{author}{\bibinfo{person}{Yizhou Sun} {and} \bibinfo{person}{Jiawei
  Han}.} \bibinfo{year}{2012}\natexlab{}.
\newblock \showarticletitle{Mining heterogeneous information networks:
  principles and methodologies}.
\newblock \bibinfo{journal}{\emph{Synthesis Lectures on Data Mining and
  Knowledge Discovery}} \bibinfo{volume}{3}, \bibinfo{number}{2}
  (\bibinfo{year}{2012}), \bibinfo{pages}{1--159}.
\newblock


\bibitem[\protect\citeauthoryear{Thackeray, Neiger, Smith, and
  Van~Wagenen}{Thackeray et~al\mbox{.}}{2012}]%
        {thackeray2012adoption}
\bibfield{author}{\bibinfo{person}{Rosemary Thackeray}, \bibinfo{person}{Brad~L
  Neiger}, \bibinfo{person}{Amanda~K Smith}, {and} \bibinfo{person}{Sarah~B
  Van~Wagenen}.} \bibinfo{year}{2012}\natexlab{}.
\newblock \showarticletitle{Adoption and use of social media among public
  health departments}.
\newblock \bibinfo{journal}{\emph{BMC public health}} \bibinfo{volume}{12},
  \bibinfo{number}{1} (\bibinfo{year}{2012}), \bibinfo{pages}{1--6}.
\newblock


\bibitem[\protect\citeauthoryear{Tibshirani}{Tibshirani}{1996}]%
        {tibshirani1996regression}
\bibfield{author}{\bibinfo{person}{Robert Tibshirani}.}
  \bibinfo{year}{1996}\natexlab{}.
\newblock \showarticletitle{Regression shrinkage and selection via the lasso}.
\newblock \bibinfo{journal}{\emph{Journal of the Royal Statistical Society:
  Series B (Methodological)}} \bibinfo{volume}{58}, \bibinfo{number}{1}
  (\bibinfo{year}{1996}), \bibinfo{pages}{267--288}.
\newblock


\bibitem[\protect\citeauthoryear{Yang and Luo}{Yang and Luo}{2017}]%
        {yang2017tracking}
\bibfield{author}{\bibinfo{person}{Xitong Yang} {and} \bibinfo{person}{Jiebo
  Luo}.} \bibinfo{year}{2017}\natexlab{}.
\newblock \showarticletitle{Tracking illicit drug dealing and abuse on
  Instagram using multimodal analysis}.
\newblock \bibinfo{journal}{\emph{ACM Transactions on Intelligent Systems and
  Technology (TIST)}} \bibinfo{volume}{8}, \bibinfo{number}{4}
  (\bibinfo{year}{2017}), \bibinfo{pages}{1--15}.
\newblock


\bibitem[\protect\citeauthoryear{Zafarani, Abbasi, and Liu}{Zafarani
  et~al\mbox{.}}{2014}]%
        {zafarani2014social}
\bibfield{author}{\bibinfo{person}{Reza Zafarani},
  \bibinfo{person}{Mohammad~Ali Abbasi}, {and} \bibinfo{person}{Huan Liu}.}
  \bibinfo{year}{2014}\natexlab{}.
\newblock \bibinfo{booktitle}{\emph{Social media mining: an introduction}}.
\newblock \bibinfo{publisher}{Cambridge University Press}.
\newblock


\bibitem[\protect\citeauthoryear{Zhang}{Zhang}{2010}]%
        {zhang2010multi}
\bibfield{author}{\bibinfo{person}{Jixian Zhang}.}
  \bibinfo{year}{2010}\natexlab{}.
\newblock \showarticletitle{Multi-source remote sensing data fusion: status and
  trends}.
\newblock \bibinfo{journal}{\emph{International Journal of Image and Data
  Fusion}} \bibinfo{volume}{1}, \bibinfo{number}{1} (\bibinfo{year}{2010}),
  \bibinfo{pages}{5--24}.
\newblock


\bibitem[\protect\citeauthoryear{Zhang, Fan, Song, Hou, Ye, Li, Zhao, Shi,
  Wang, and Xiong}{Zhang et~al\mbox{.}}{2019}]%
        {zhang2019your}
\bibfield{author}{\bibinfo{person}{Yiming Zhang}, \bibinfo{person}{Yujie Fan},
  \bibinfo{person}{Wei Song}, \bibinfo{person}{Shifu Hou},
  \bibinfo{person}{Yanfang Ye}, \bibinfo{person}{Xin Li},
  \bibinfo{person}{Liang Zhao}, \bibinfo{person}{Chuan Shi},
  \bibinfo{person}{Jiabin Wang}, {and} \bibinfo{person}{Qi Xiong}.}
  \bibinfo{year}{2019}\natexlab{}.
\newblock \showarticletitle{Your style your identity: Leveraging writing and
  photography styles for drug trafficker identification in darknet markets over
  attributed heterogeneous information network}. In
  \bibinfo{booktitle}{\emph{The World Wide Web Conference}}.
  \bibinfo{pages}{3448--3454}.
\newblock


\bibitem[\protect\citeauthoryear{Zhang and Sabuncu}{Zhang and Sabuncu}{2018}]%
        {zhang2018generalized}
\bibfield{author}{\bibinfo{person}{Zhilu Zhang} {and} \bibinfo{person}{Mert~R
  Sabuncu}.} \bibinfo{year}{2018}\natexlab{}.
\newblock \showarticletitle{Generalized cross entropy loss for training deep
  neural networks with noisy labels}.
\newblock \bibinfo{journal}{\emph{arXiv preprint arXiv:1805.07836}}
  (\bibinfo{year}{2018}).
\newblock


\bibitem[\protect\citeauthoryear{Zhao, Skums, Zelikovsky, Sevigny, Swahn,
  Strasser, Huang, and Wu}{Zhao et~al\mbox{.}}{2020}]%
        {zhao2020computational}
\bibfield{author}{\bibinfo{person}{Fengpan Zhao}, \bibinfo{person}{Pavel
  Skums}, \bibinfo{person}{Alexander Zelikovsky}, \bibinfo{person}{Eric~L
  Sevigny}, \bibinfo{person}{Monica~Haavisto Swahn}, \bibinfo{person}{Sheryl~M
  Strasser}, \bibinfo{person}{Yan Huang}, {and} \bibinfo{person}{Yubao Wu}.}
  \bibinfo{year}{2020}\natexlab{}.
\newblock \showarticletitle{Computational approaches to detect illicit drug ads
  and find vendor communities within social media platforms}.
\newblock \bibinfo{journal}{\emph{IEEE/ACM transactions on computational
  biology and bioinformatics}} (\bibinfo{year}{2020}).
\newblock


\bibitem[\protect\citeauthoryear{Zheng, Zhu, Li, Pang, Wang, and Li}{Zheng
  et~al\mbox{.}}{2020}]%
        {zheng2020feature}
\bibfield{author}{\bibinfo{person}{Qinghai Zheng}, \bibinfo{person}{Jihua Zhu},
  \bibinfo{person}{Zhongyu Li}, \bibinfo{person}{Shanmin Pang},
  \bibinfo{person}{Jun Wang}, {and} \bibinfo{person}{Yaochen Li}.}
  \bibinfo{year}{2020}\natexlab{}.
\newblock \showarticletitle{Feature concatenation multi-view subspace
  clustering}.
\newblock \bibinfo{journal}{\emph{Neurocomputing}}  \bibinfo{volume}{379}
  (\bibinfo{year}{2020}), \bibinfo{pages}{89--102}.
\newblock


\bibitem[\protect\citeauthoryear{Zhou, Sun, Liu, and Lau}{Zhou
  et~al\mbox{.}}{2015}]%
        {zhou2015c}
\bibfield{author}{\bibinfo{person}{Chunting Zhou}, \bibinfo{person}{Chonglin
  Sun}, \bibinfo{person}{Zhiyuan Liu}, {and} \bibinfo{person}{Francis Lau}.}
  \bibinfo{year}{2015}\natexlab{}.
\newblock \showarticletitle{A C-LSTM neural network for text classification}.
\newblock \bibinfo{journal}{\emph{arXiv preprint arXiv:1511.08630}}
  (\bibinfo{year}{2015}).
\newblock


\bibitem[\protect\citeauthoryear{Zhou, Qi, Zheng, Xu, Bao, and Xu}{Zhou
  et~al\mbox{.}}{2016a}]%
        {zhou2016text}
\bibfield{author}{\bibinfo{person}{Peng Zhou}, \bibinfo{person}{Zhenyu Qi},
  \bibinfo{person}{Suncong Zheng}, \bibinfo{person}{Jiaming Xu},
  \bibinfo{person}{Hongyun Bao}, {and} \bibinfo{person}{Bo Xu}.}
  \bibinfo{year}{2016}\natexlab{a}.
\newblock \showarticletitle{Text classification improved by integrating
  bidirectional LSTM with two-dimensional max pooling}.
\newblock \bibinfo{journal}{\emph{arXiv preprint arXiv:1611.06639}}
  (\bibinfo{year}{2016}).
\newblock


\bibitem[\protect\citeauthoryear{Zhou, Sani, Lee, and Luo}{Zhou
  et~al\mbox{.}}{2016c}]%
        {zhou2016understanding}
\bibfield{author}{\bibinfo{person}{Yiheng Zhou}, \bibinfo{person}{Numair Sani},
  \bibinfo{person}{Chia-Kuei Lee}, {and} \bibinfo{person}{Jiebo Luo}.}
  \bibinfo{year}{2016}\natexlab{c}.
\newblock \showarticletitle{Understanding illicit drug use behaviors by mining
  social media}.
\newblock \bibinfo{journal}{\emph{arXiv preprint arXiv:1604.07096}}
  (\bibinfo{year}{2016}).
\newblock


\bibitem[\protect\citeauthoryear{Zhou, Sani, and Luo}{Zhou
  et~al\mbox{.}}{2016b}]%
        {zhou2016fine}
\bibfield{author}{\bibinfo{person}{Yiheng Zhou}, \bibinfo{person}{Numair Sani},
  {and} \bibinfo{person}{Jiebo Luo}.} \bibinfo{year}{2016}\natexlab{b}.
\newblock \showarticletitle{Fine-grained mining of illicit drug use patterns
  using social multimedia data from Instagram}. In
  \bibinfo{booktitle}{\emph{2016 IEEE International Conference on Big Data (Big
  Data)}}. IEEE, \bibinfo{pages}{1921--1930}.
\newblock


\end{thebibliography}
\end{document}